\documentclass[a4paper,11pt]{article}

\usepackage{a4wide}
\usepackage{amssymb,amsmath,amsthm}
\usepackage{booktabs,multirow}
\usepackage{graphicx,subcaption}
\usepackage[english]{babel}
\usepackage[round]{natbib}
\usepackage[hyphens]{url}
\usepackage{hyperref}
\hypersetup{breaklinks=true}
\newtheorem{remark}{Remark}

\numberwithin{equation}{section}
\graphicspath{{Figures/}}

\begin{document}

\title{SciSports: Learning football kinematics through two-dimensional tracking data}

\author{Anatoliy Babic, Harshit Bansal, Gianluca Finocchio, \\Julian Golak, Mark Peletier, Jim Portegies, \\Clara Stegehuis, Anuj Tyagi, Roland Vincze, \\William Weimin Yoo}

\maketitle

\begin{abstract}

SciSports is a Dutch startup company specializing in football analytics. This paper describes a joint research effort with SciSports,  during the Study Group Mathematics with Industry 2018 at Eindhoven, the Netherlands. The main challenge that we addressed was to automatically process empirical football players' trajectories, in order to extract useful information from them.

The data provided to us was two-dimensional positional data during entire matches. We developed methods based on Newtonian mechanics and the Kalman filter, Generative Adversarial Nets and Variational Autoencoders. In addition, we trained a discriminator network to recognize and discern different movement patterns of players. 

The Kalman-filter approach yields an interpretable model, in which a small number of player-dependent parameters can be fit; in theory this could be used to distinguish among players. 

The Generative-Adversarial-Nets approach appears promising in theory, and some initial tests showed an improvement with respect to the baseline, but the limits in time and computational power meant that we could not fully explore it. 
We also trained a Discriminator network to distinguish between two players  based on their trajectories; after training, the network managed to distinguish between some pairs of players, but not between others. After training, the Variational Autoencoders generated trajectories that are difficult to distinguish, visually, from the data.

These experiments provide an indication that deep generative models can learn the underlying structure and statistics of football players' trajectories. This can serve as a starting point for determining player qualities based on such trajectory data. 

\noindent

\end{abstract}


\section{Introduction}\label{sci-sec:introduction}
SciSports (\url{http://www.scisports.com/}) is a Dutch sports analytics company taking a data-driven approach to football. The company conducts scouting activities for football clubs, gives advice to football players about which football club might suit them best, and quantifies the abilities of football players through various performance metrics. 
So far, most of these activities have been supported by either coarse event data, such as line-ups and outcomes of matches, or more fine-grained event data such as completed passes, distances covered by players, yellow cards received and goals scored. 

In the long term,  SciSports aims to install specialized cameras and sensors across  football fields to create a two- and three-dimensional virtual rendering of the matches, by recording players' coordinate positions and gait data in millisecond time intervals. From this massive amount of data, SciSports is interested in predicting future game courses and extracting useful analytics. Insights gained from this learning process can be used as preliminary steps towards determining the quality and playing style of football players. In this project  we based our work on a dataset containing the recorded two-dimensional positions of all players and the ball during 14 standard football matches at $0.1$ second time intervals. 

Football kinematics such as acceleration, maximal sprinting speed and distance covered during a match can be extracted automatically from trajectory data. However, there are also important unobservable factors/features determining the soccer game, e.g., a player can be of enormous value to a game without being anywhere near the ball. These latent factors are key to understanding the drivers of motion and their roles in predicting future game states. There are in general two basic approaches to uncovering these factors:  we can either postulate a model or structure for these factors, based on physical laws and other domain knowledge (model-based), or we can use machine learning techniques and let the algorithms discover these factors on their own (data-driven). 

Model-based approaches have been widely used to analyze football trajectories. Examples in the literature include statistical models such as state space models \cite{sci-bib:yu2003a,sci-bib:yu2003b,sci-bib:ren2008} and physical models based on equations of motion and aerodynamics \cite{sci-bib:goff2009}. These methods have the advantage of producing interpretable results and they can quickly give reasonable predictions using relatively few past observations. In Section~\ref{sci-sec:kalman}, we build state space models based on principles of Newtonian mechanics to illustrate these approaches.

The need to specify an explicit model is a drawback, however, since human players probably follow  complicated rules of behavior. To this end, data-driven approaches embody the promise of taking advantage of having large amounts of data  through  machine learning algorithms, without specifying the model;  in a sense the model is chosen by the algorithm as part of the training. 

We implemented a Variational Autoencoder (VAE), as introduced by \cite{sci-bib:vae}, and a Generative Adversarial Net (GAN) as developed in \cite{sci-bib:Goodfellow2014}.

The paper is organized as follows. In the next section, we describe the two-dimensional positional data used for our analyses. We present the model-based state-space approach in Section \ref{sci-sec:methods} and the data-driven methods based on GANs and VAEs in Sections \ref{sci-sec:gan} and \ref{sci-sec:vae}, respectively. We introduce the discriminator network to differentiate movements in \ref{sci-sec:dis}. We conclude in Section~\ref{sci-sec:con} and discuss future work. 

\medskip

The \texttt{R} and \texttt{Python} codes used to reproduce all our analyses can be found in \url{https://bitbucket.org/AnatoliyBabic/swi-scisports-2018}.

\begin{figure}[t]
	\centerline{\includegraphics[scale=0.42]{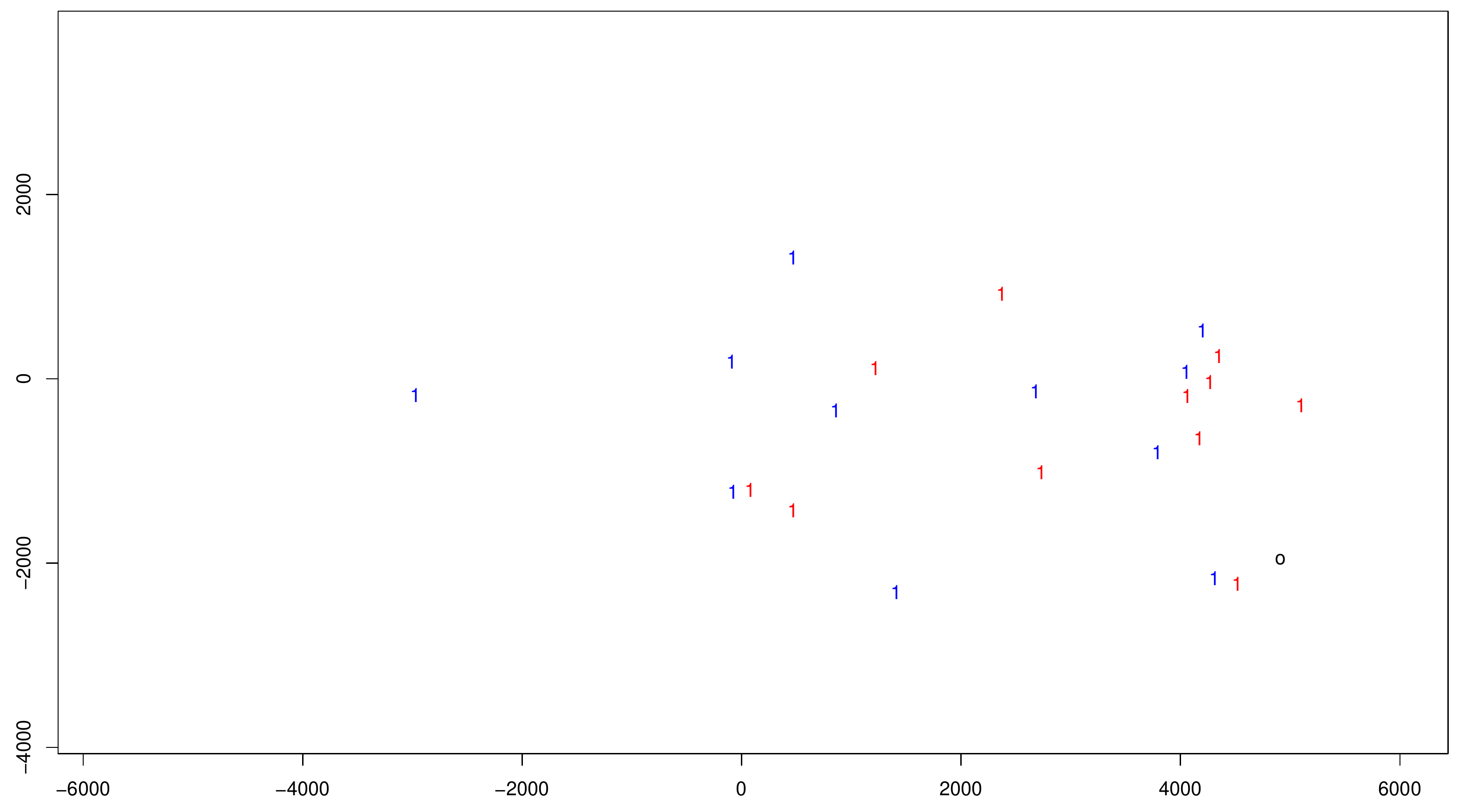}}
	\caption{A snapshot in time ($\approx$ 2 minutes into the game) of the positional data for all players (blue and red teams) and the ball (circle). Note that the goalkeepers can be identified as the players standing at the leftmost and rightmost positions on the field.}
	\label{sci-fig:game}
\end{figure}

\section{The data}\label{sci-sec:data}
The data that we used for this project was provided by SciSports and is taken from $14$ complete $90$-minute football matches. For each player and ball ($23$ entities total) the $(x,y)$-coordinates on the field have been recorded  with a resolution of $10$ cm and $10$ frames per second; i.e., the trajectory of a player on a $10$ seconds timespan corresponds to a $(2\times 100)$-vector of $(x,y)$-coordinates. The field measures $68$ by $105$ meters, and the origin of the coordinate system is the center of the pitch. For all football fields illustrated in this report, the dimensions are given in centimeters, which means that the field corresponds to the rectangle $[-5250,5250]\times [-3400,3400]$. 

For illustration, Figure \ref{sci-fig:game} shows a single-time snapshot of the positional data for the ball and all players. 

\section{Methods: model-based}\label{sci-sec:methods}
In this section we describe a model-based approach to extract information from the data. With this approach we have two goals: first, to extract velocities from the position data in such a way that the impact of the noise in position measurements is minimized, and secondly, to estimate acceleration profiles of different players.

\subsection{Newtonian mechanics and the Kalman filter}\label{sci-sec:kalman}
\subsection*{A single football player}
We first consider the case of modeling the movement of one football player in the first match. 
We assume that this player is not a goalkeeper, since we would like to model movement ranges that span at least half the field. The data provides a player's $(x,y)$-position at every fixed $100$ milliseconds as long as he remains in the game. Let $\Delta t$ be the time difference between successive timesteps, and let us denote a player's position in the $(x,y)$ plane at timestep $t$ as $\boldsymbol{x}_t$, with the velocity and acceleration as $\boldsymbol{v}_t$ and $\boldsymbol{a}_t$; they are related by $\boldsymbol{a}_t=d\boldsymbol{v}_t/dt$ and $\boldsymbol{v}_t=d\boldsymbol{x}_t/dt$. By approximating these derivatives by finite differences  we obtain
\begin{align}\label{sci-eq:newton}
	\boldsymbol{x}_t&=\boldsymbol{x}_{t-1}+\Delta t\,\boldsymbol{v}_{t-1}+\frac{1}{2}(\Delta t)^2\boldsymbol{a}_t,\nonumber\\
	\boldsymbol{v}_t&=\boldsymbol{v}_{t-1}+\Delta t\, \boldsymbol{a}_t.
\end{align}

We now model the acceleration $\boldsymbol a_t$. We assume that at each timestep $t$ the acceleration $\boldsymbol a_t$ is independently and normally  distributed with mean $\boldsymbol{0}$ and unknown covariance matrix $\boldsymbol{Q}$ (we write this as $\boldsymbol a_t \sim \mathrm N(\boldsymbol 0,\boldsymbol Q)$). Since acceleration is proportional to force by Newton's second law of motion, this induces a normal distribution on the corresponding force exerted by the player, and the exponential decay of its tails translate to natural limits imposed on muscular work output.   

In view of \eqref{sci-eq:newton}, we take position and velocity $(\boldsymbol{x}_t,\boldsymbol{v}_t)$ as our underlying state vector, and we consider the following model:
\begin{align}
	\label{sci-eq:latent}
	\begin{pmatrix}\boldsymbol{x}_t\\ \boldsymbol{v}_t\end{pmatrix}
	&=\underbrace{\begin{pmatrix}\boldsymbol{I}_2&\Delta t\boldsymbol{I}_2\\ \boldsymbol{0}&\boldsymbol{I}_2\end{pmatrix}}_{\boldsymbol{T}_t}\begin{pmatrix}\boldsymbol{x}_{t-1}\\ \boldsymbol{v}_{t-1}\end{pmatrix}
	+\underbrace{\begin{pmatrix}\frac{1}{2}(\Delta t)^2\boldsymbol{I}_2\\ \Delta t\boldsymbol{I}_2\end{pmatrix}}_{\boldsymbol{R}_t}\boldsymbol{a}_t,\\
	\boldsymbol{\eta}_t&=
	\underbrace{\begin{pmatrix}
			1&0&0&0\\
			0&1&0&0
	\end{pmatrix}}_{\boldsymbol{W}_t}
	\underbrace{\begin{pmatrix}
			\boldsymbol{x}_t\\ 
			\boldsymbol{v}_t
	\end{pmatrix}}_{\boldsymbol{z}_t}
	+
	\;\boldsymbol{\varepsilon}_t,
	\label{sci-eq:observe}
\end{align}
In the state equation \eqref{sci-eq:latent}, the state vector $\boldsymbol{z}_t:=(\boldsymbol{x}_t,\boldsymbol{v}_t)$ propagates forward in time according to the Newtonian dynamics of \eqref{sci-eq:newton}, driven by an acceleration $\boldsymbol{a}_t\sim\mathrm{N}(\boldsymbol{0},\boldsymbol{Q})$.
In the observation equation \eqref{sci-eq:observe}, the observed quantity~$\boldsymbol \eta_t$ records the player's position and not his/her velocity, and we assume that these position data are recorded with Gaussian measurement errors: $\boldsymbol{\varepsilon}_t\sim\mathrm{N}(\boldsymbol{0},\boldsymbol{\Sigma})$ with $\boldsymbol{\Sigma}=\mathrm{Diag}(\sigma_x^2,\sigma_y^2)$. 
We initialize $\boldsymbol{z}_1\sim\mathrm{N}(\boldsymbol{0},\boldsymbol{P}_1)$ and we assume that $\boldsymbol{\varepsilon}_t,\boldsymbol{a}_t$, and $\boldsymbol{z}_1$ are mutually independent, and independent across different times. 

\medskip

We use a Kalman filter to integrate this model with the measurements; this should lead to an estimate for the velocity that is less noisy than simply calculating finite differences. However, the Kalman filter parameters depend on the noise levels as characterized by the player's acceleration variance $\boldsymbol Q$ and the measurement error parameters $\sigma_x,\sigma_y$, and these we do not  know; therefore we combine the Kalman filter with parameter estimation.

\medskip

In each Kalman-filter timestep we assume that we have access to observations~$\boldsymbol\eta_t$, and we compute the one-step state prediction $\boldsymbol{Z}_{t+1}=\mathrm{E}(\boldsymbol{z}_{t+1}|\boldsymbol{\eta}_t,\dotsc,\boldsymbol{\eta}_1)$ and its error $\boldsymbol{\delta}_t=\boldsymbol{\eta}_t-\boldsymbol{W}_t\boldsymbol{Z}_t$, in conjunction with their estimated covariance matrices $\boldsymbol{P}_{t+1}=\mathrm{Var}(\boldsymbol{z}_{t+1}|\boldsymbol{\eta}_t,\dotsc,\boldsymbol{\eta}_1)$ and $\boldsymbol{F}_t=\mathrm{Var}(\boldsymbol{\delta}_t)=\boldsymbol{W}_t\boldsymbol{P}_t\boldsymbol{W}_t^T+\boldsymbol{\Sigma}$. The Kalman recursion formulas for these calculations are given by (see Appendix A of \citealp{sci-bib:kfas})
\begin{subequations}
	\label{sci-eq:Kalman}
	\begin{align}
		\boldsymbol{Z}_{t+1}&=\boldsymbol{T}_t(\boldsymbol{Z}_t+\boldsymbol{K}_t\boldsymbol{F}_t^{-1}\boldsymbol{\delta}_t)\\
		\boldsymbol{P}_{t+1}&=\boldsymbol{T}_t(\boldsymbol{P}_t-\boldsymbol{K}_t\boldsymbol{F}_t^{-1}\boldsymbol{K}_t^T)\boldsymbol{T}_t^T+\boldsymbol{R}_t\boldsymbol{Q}\boldsymbol{R}_t^T,
	\end{align}
\end{subequations}
where $\boldsymbol{K}_t=\boldsymbol{P}_t\boldsymbol{W}_t^T$. For given values of $\boldsymbol Q$ and $\sigma_x,\sigma_y$ this leads to time courses of the state $\boldsymbol Z_t$, the covariance $\boldsymbol P_t$, and the derived quantities $\boldsymbol \delta_t$ and $\boldsymbol F_t$.

We have a total of $6$ unknown parameters in our state space model, i.e., the two diagonal entries of $\boldsymbol{\Sigma}$ and all the $2\times2$ entries of $\boldsymbol{Q}$ (we did not exploit the symmetry of $\boldsymbol Q$). 
Given the result of a calculation for given $\boldsymbol Q$ and $\sigma_x,\sigma_y$, the log-likelihood function~\citep{sci-bib:kfas} is given by
\begin{align}
	\label{sci-eq:log-likelihood-Kalman}
	l_n=-\frac{np}{2}\log{(2\pi)}-\frac{1}{2}\sum_{t=1}^n\left(\log{\det{\boldsymbol{F}_t}}+\boldsymbol{\delta}_t^T\boldsymbol{F}_t^{-1}\boldsymbol{\delta}_t\right),
\end{align}
where $p$ is the dimension of $\boldsymbol{\eta}_t$ at a fixed $t$, which in our present case is $2$. We then compute the maximum likelihood estimator for the $6$ covariance parameters using the Broyden-Fletcher-Goldfarb-Shanno (BFGS) optimization algorithm.

This setup leads to the following multilevel iteration. 
\begin{enumerate}
	\item We select the first 10 timesteps from the data; this means that we know the values of $\boldsymbol \eta_1$ to $\boldsymbol \eta_{10}$.
	\item \label{sci-en:iteration-Kalman}
	At the outer level we maximize the log-likelihood function~\eqref{sci-eq:log-likelihood-Kalman} with respect to~$\boldsymbol Q$ and  $\sigma_x,\sigma_y$.
	\item At the inner level, i.e.\ for each evaluation of the log-likelihood, we run the Kalman filter~\eqref{sci-eq:Kalman} for 10 steps, ending at time $t=11$. 
	\item After completing the optimization over $\boldsymbol Q$ and  $\sigma_x,\sigma_y$ for this choice of 10 timesteps, we have both an estimate of $\boldsymbol Q$ and  $\sigma_x,\sigma_y$ during that period and a prediction for $\boldsymbol z_t = (\boldsymbol x_t,\boldsymbol v_t)$, for $t=1,\dots,11$. We then shift the 10-step window by one timestep, to $2,\dots,11$, and go back to step~\ref{sci-en:iteration-Kalman}. 
\end{enumerate}
At the end of this process, we have for each 10-step window of times a series of estimates of $\boldsymbol x_t$, $\boldsymbol v_t$, $\boldsymbol P_t$, $\boldsymbol Q$, and $\sigma_x,\sigma_y$. 

\begin{remark}
	Each of the 11-step runs of the Kalman filter equations~\eqref{sci-eq:Kalman} needs to be initialized. We initialize $\boldsymbol z_1$ randomly, drawn from $\mathrm N(\boldsymbol 0, \boldsymbol P_1)$, as mentioned above. Concerning the choice of $\boldsymbol{P}_1$, a commonly used default is to set $\boldsymbol{P}_1=10^7\boldsymbol{I}$ as a diffuse prior distribution. However, this is numerically unstable and prone to cumulative roundoff errors. Instead, we use the exact diffuse initialization method by decomposing $\boldsymbol{P}_1$ into its diffusive and non-diffusive parts; for more details see \cite{sci-bib:koopman2003}.
\end{remark}

\begin{remark}
	In actual implementation, some technical modifications are needed to speed up computations, particularly when $\boldsymbol{\eta}_t$ consists of high-dimensional observations at each time point (which happens when we estimate all 23 entities, as we do below). To solve for this dimensionality issue and to avoid direct inversion of $\boldsymbol{F}_t$, the state space model of \eqref{sci-eq:observe} and \eqref{sci-eq:latent} is recast into an equivalent univariate form and the latent states are estimated using a univariate Kalman filter (cf.~\citealp{sci-bib:koopman2000}).
\end{remark}

The Kalman filter algorithm and parameter estimation (including the univariate formulation and diffuse initialization) were performed using the \texttt{KFAS} package (see \citealp{sci-bib:kfas}) in the \texttt{R} software package.

\subsection*{Results for a single player}
We modeled the movement of the player with number 3, who appears to  hold the position of left central midfielder, and who was in the pitch for the entire game.  As described above, we use a sliding window of $10$ training samples for predictions, such that we first use $10$ time points to predict the $11$th point (one-step-ahead), then we shift the window one timestep ahead and use the next $10$ time points to predict the $12$th point and so on.

\begin{figure}[ht!]
	\centerline{\includegraphics[scale=0.43]{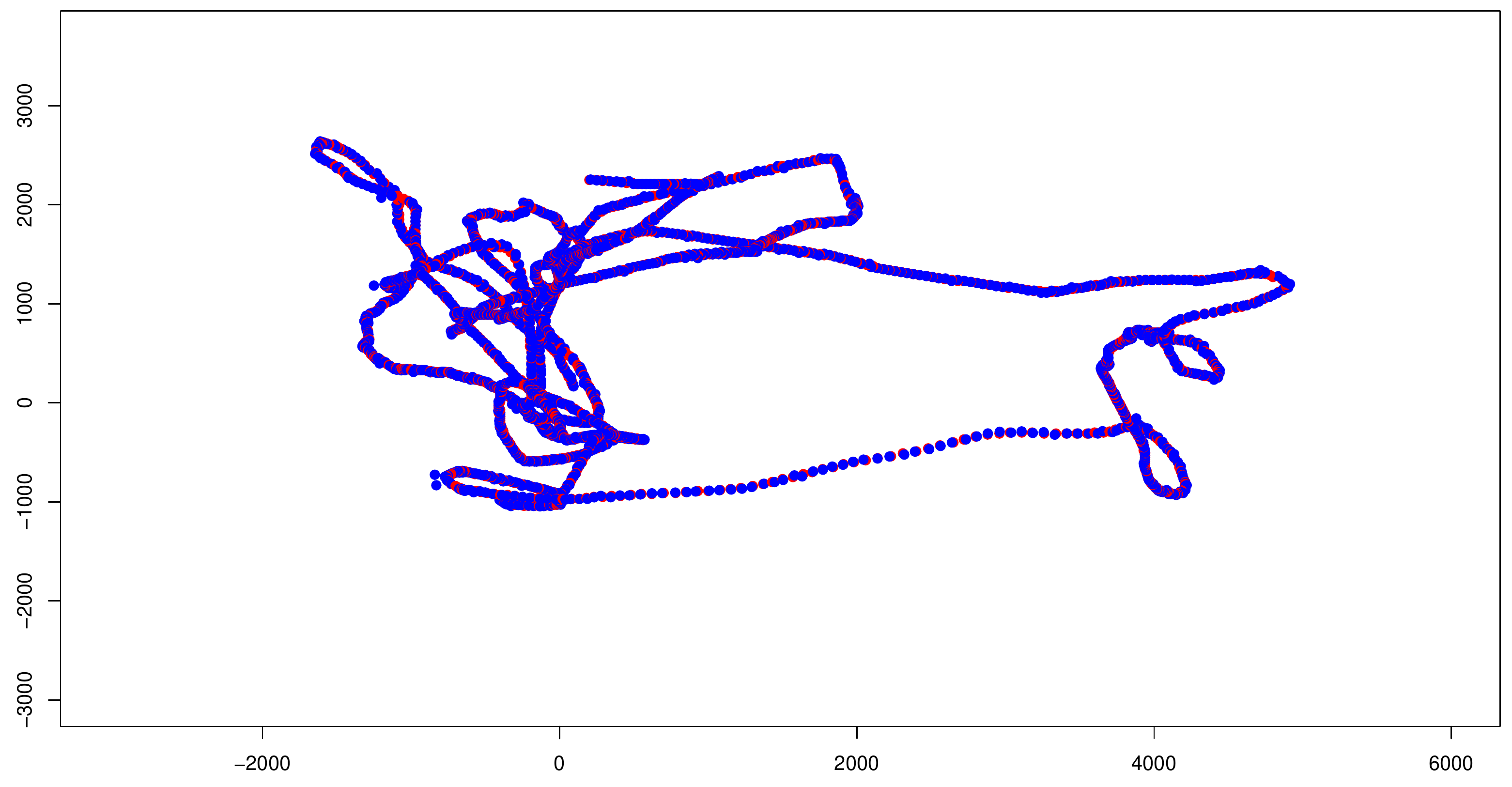}}
	\caption{Blue: One-step-ahead predicted position, Red: True recorded position.}
	\label{sci-fig:prediction}
\end{figure}

\begin{figure}[ht!]
	\centerline{\includegraphics[scale=0.45]{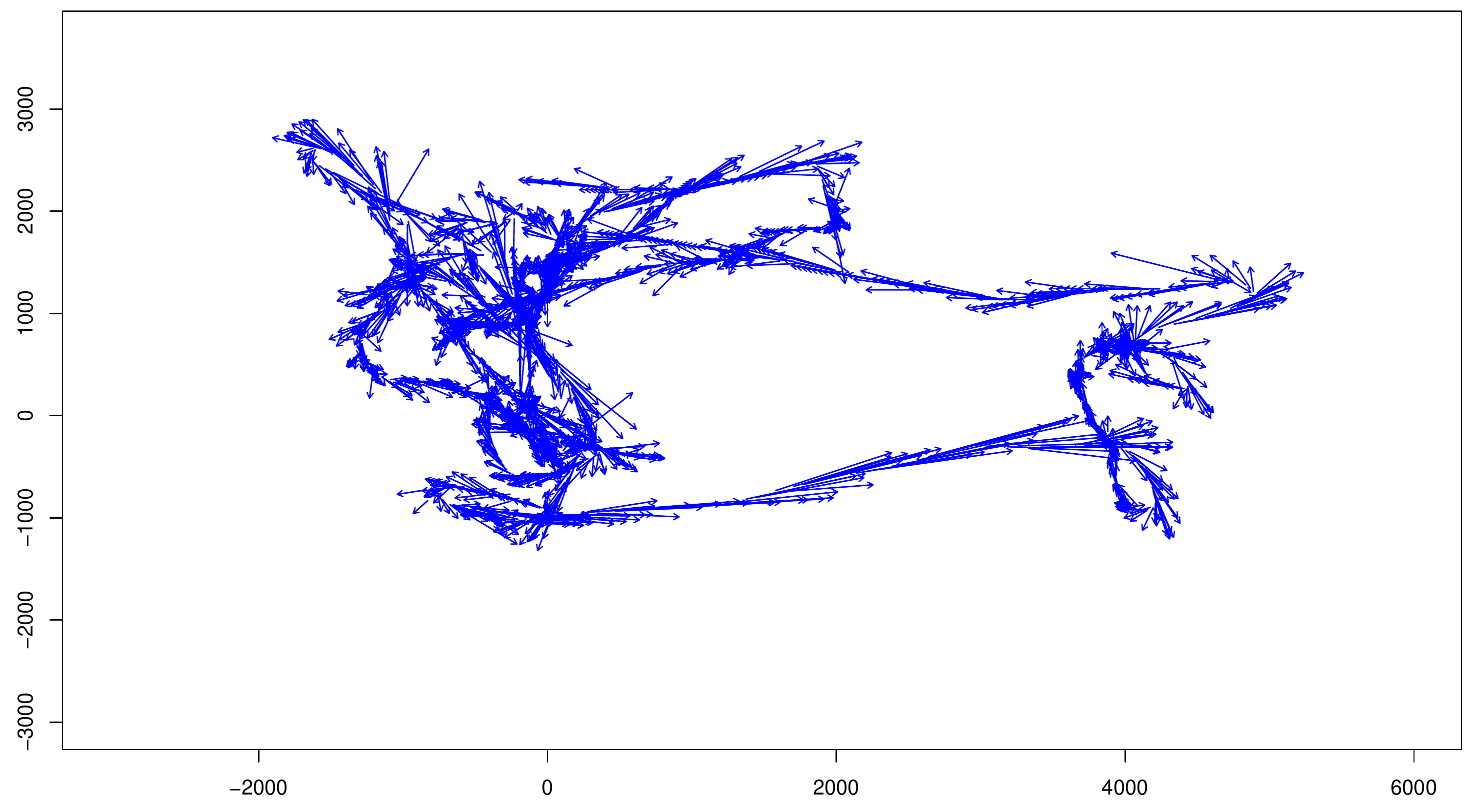}}
	\caption{One-step-ahead predicted velocity vector field $\boldsymbol{v}_t$, arrow points to direction of motion and vector length is speed.}
	\label{sci-fig:velocity}
\end{figure}

Figure \ref{sci-fig:prediction} shows one-step-ahead predicted positions of our midfielder (blue dots) for the first 2500 time points. We see that the state space model is able to make accurate predictions (when compared to the red true positions), even if we have used only the past $10$ locations in our algorithm. Moreover, the model is able to trace out complicated movements and sharp corners as is evident from the figure. 

\begin{figure}[ht!]
	\centerline{\includegraphics[scale=0.3]{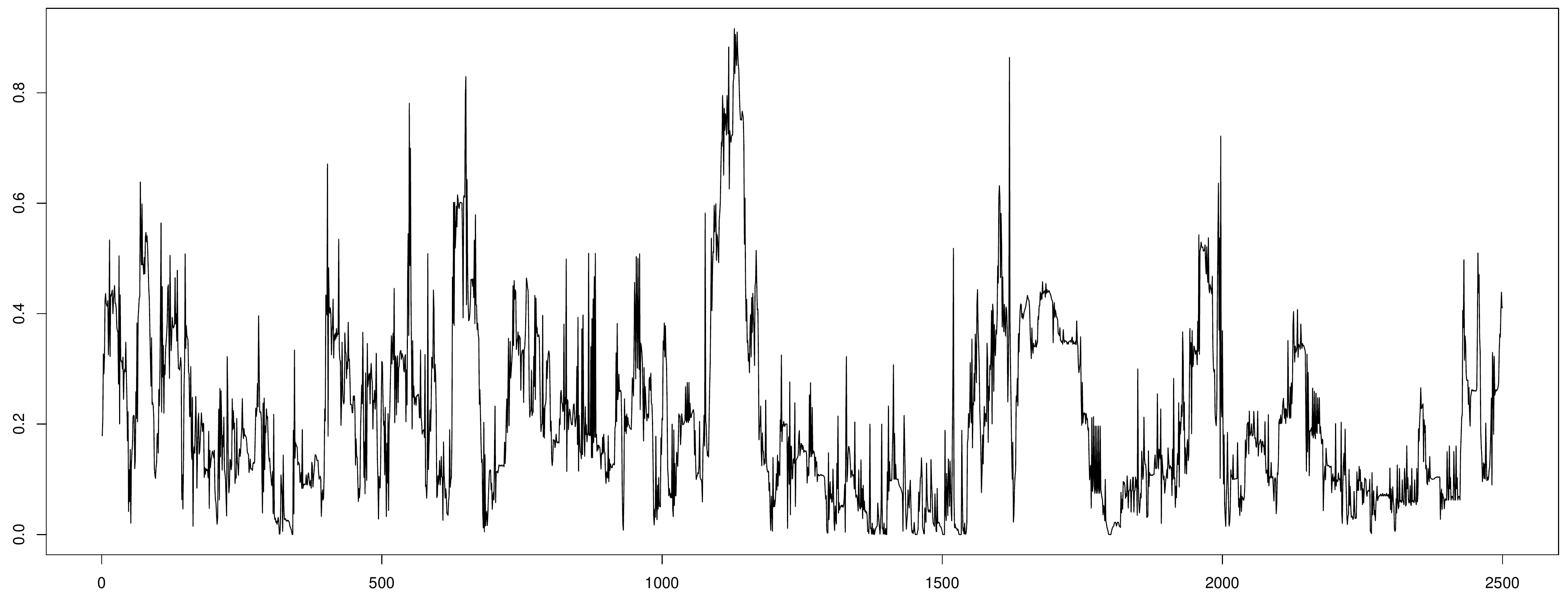}}
	\caption{One-step-ahead predicted speed $\|\boldsymbol{v}_t\|$ ($y$-axis) against timesteps ($x$-axis).}
	\label{sci-fig:speed}
\end{figure}

As mentioned above, one reason for applying a Kalman filter to the data is to extract the velocity. Figure \ref{sci-fig:velocity} illustrates the velocity vectors as arrows tangent to the position curve. 
We also plot the scalar speeds $\|\boldsymbol{v}_t\|$ against the 2500 time points in Figure~\ref{sci-fig:speed}.

To see the correspondence between these three figures, let us focus on a distinguishing stretch of movement made by our midfielder, who starts at $(0, -1000)$, then sprints towards the goal post in the East, make two loops towards the North and again moved back across the field to the West, thus making a somewhat elongated rectangle on the field. We know that he is sprinting to the goal from Figure \ref{sci-fig:velocity} due to the long arrows pointing to the East, with exact magnitudes given by the peak slightly after time $1000$ in Figure \ref{sci-fig:speed}. The midfielder has relatively lower speeds when making the double loop (from time $1200$ to $1500$ in Figure \ref{sci-fig:speed}) and then he picks up the momentum when moving towards the West, as is evident from the marked increase in speeds after time $1500$.

\begin{figure}[ht!]
	\centerline{\includegraphics[scale=0.45]{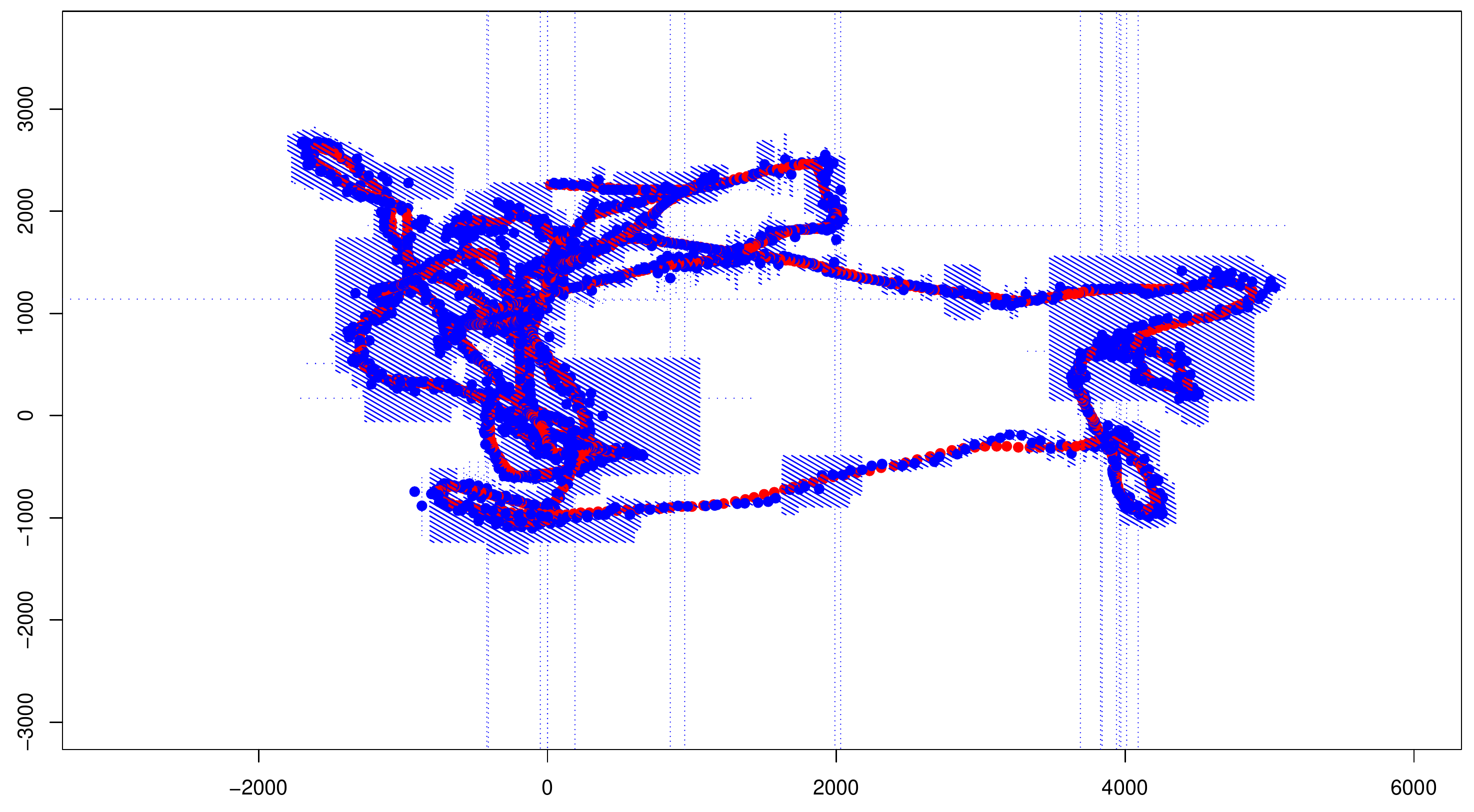}}
	\caption{Blue dot: 5-step-ahead predicted position; blue square: $95\%$-prediction rectangle; red dot: true recorded position. The horizontal and vertical lines are artefacts of the algorithm.}
	\label{sci-fig:prediction5}
\end{figure}

Figure \ref{sci-fig:prediction5} shows the predictive performance of this model for longer time horizons; in this case we are using $10$ time points to predict $5$ steps ahead. When compared with the one-step-ahead case of Figure \ref{sci-fig:prediction}, we see that there is some deterioration in this model's predictive capability, particularly for places where the player's trajectory is curved.  From this plot, we can deduce that positional uncertainties are the greatest when the midfielder is moving in loops or in circles. 

\subsection*{Results for the ball and all $22$ football players}
Let us now consider the general case of modeling all $22$ football players, including goalkeepers, and the ball (collectively called `entities'). A snapshot of the positional data at around $2$ minutes into the game is shown in Figure \ref{sci-fig:game}. We choose the same equations for all entities, giving for all $k=1,\dotsc,23$,
\begin{align}\label{sci-eq:newtonp}
	\boldsymbol{x}_t^{(k)}&=\boldsymbol{x}_{t-1}^{(k)}+\Delta t\,\boldsymbol{v}_{t-1}^{(k)}+\frac{1}{2}(\Delta t)^2\boldsymbol{a}_t^{(k)},\nonumber\\
	\boldsymbol{v}_t^{(k)}&=\boldsymbol{v}_{t-1}^{(k)}+\Delta t\,\boldsymbol{a}_t^{(k)}.
\end{align}

By stacking up $23$ copies of the single player case \eqref{sci-eq:observe} and \eqref{sci-eq:latent}, we convert the equations of motion above to the following state space model:
\begin{align*}
	\begin{pmatrix}
		\boldsymbol{x}_t^{(1)}\\ \boldsymbol{v}_t^{(1)}\\ \boldsymbol{x}_t^{(2)}\\ \boldsymbol{v}_t^{(2)}\\ \vdots\\ \boldsymbol{x}_t^{(23)}\\ \boldsymbol{v}_t^{(23)}\end{pmatrix}
	&=\begin{pmatrix}
		\boldsymbol{I}_2&\Delta t\boldsymbol{I}_2&\boldsymbol{0}&\boldsymbol{0}&\cdots&\boldsymbol{0}&\boldsymbol{0}\\
		\boldsymbol{0}&\boldsymbol{I}_2&\boldsymbol{0}&\boldsymbol{0}&\cdots&\boldsymbol{0}&\boldsymbol{0}\\
		\boldsymbol{0}&\boldsymbol{0}&\boldsymbol{I}_2&\Delta t\boldsymbol{I}_2&\cdots&\boldsymbol{0}&\boldsymbol{0}\\
		\boldsymbol{0}&\boldsymbol{0}&\boldsymbol{0}&\boldsymbol{I}_2&\cdots&\boldsymbol{0}&\boldsymbol{0}\\
		\vdots&\vdots&\vdots&\vdots&\ddots&\vdots&\vdots\\
		\boldsymbol{0}&\boldsymbol{0}&\boldsymbol{0}&\boldsymbol{0}&\cdots&\boldsymbol{I}_2&\Delta t\boldsymbol{I}_2\\
		\boldsymbol{0}&\boldsymbol{0}&\boldsymbol{0}&\boldsymbol{0}&\cdots&\boldsymbol{0}&\boldsymbol{I}_2\end{pmatrix}
	\begin{pmatrix}
		\boldsymbol{x}_{t-1}^{(1)}\\ \boldsymbol{v}_{t-1}^{(1)}\\ \boldsymbol{x}_{t-1}^{(2)}\\ \boldsymbol{v}_{t-1}^{(2)}\\ \vdots\\ \boldsymbol{x}_{t-1}^{(23)}\\ \boldsymbol{v}_{t-1}^{(23)}\end{pmatrix}\\
	&\qquad+\begin{pmatrix}\frac{1}{2}(\Delta t)^2\boldsymbol{I}_2&\boldsymbol{0}&\boldsymbol{0}&\cdots&\boldsymbol{0}&\boldsymbol{0}\\
		\Delta t\boldsymbol{I}_2&\boldsymbol{0}&\boldsymbol{0}&\cdots&\boldsymbol{0}&\boldsymbol{0}\\
		\boldsymbol{0}&\frac{1}{2}(\Delta t)^2\boldsymbol{I}_2&\boldsymbol{0}&\cdots&\boldsymbol{0}&\boldsymbol{0}\\
		\boldsymbol{0}&\Delta t\boldsymbol{I}_2&\boldsymbol{0}&\cdots&\boldsymbol{0}&\boldsymbol{0}\\
		\vdots&\vdots&\vdots&\ddots&\vdots&\vdots\\
		\boldsymbol{0}&\boldsymbol{0}&\boldsymbol{0}&\cdots&\boldsymbol{0}&\frac{1}{2}(\Delta t)^2\boldsymbol{I}_2\\
		\boldsymbol{0}&\boldsymbol{0}&\boldsymbol{0}&\cdots&\boldsymbol{0}&\Delta t\boldsymbol{I}_2\end{pmatrix}\begin{pmatrix}\boldsymbol{a}_t^{(1)}\\ \boldsymbol{a}_t^{(2)}\\ \vdots\\ \boldsymbol{a}_t^{(23)}\end{pmatrix},
\end{align*}
with measurement vector
\begin{align*}
	\boldsymbol{y}_t=\begin{pmatrix}
		\boldsymbol{I}_2&\boldsymbol{0}&\boldsymbol{0}&\boldsymbol{0}&\boldsymbol{0}&\boldsymbol{0}&\cdots&\boldsymbol{0}&\boldsymbol{0}\\
		\boldsymbol{0}&\boldsymbol{0}&\boldsymbol{I}_2&\boldsymbol{0}&\boldsymbol{0}&\boldsymbol{0}&\cdots&\boldsymbol{0}&\boldsymbol{0}\\
		\boldsymbol{0}&\boldsymbol{0}&\boldsymbol{0}&\boldsymbol{0}&\boldsymbol{I}_2&\boldsymbol{0}&\cdots&\boldsymbol{0}&\boldsymbol{0}\\
		\vdots&\vdots&\vdots&\vdots&\vdots&\vdots&\ddots&\vdots&\vdots\\
		\boldsymbol{0}&\boldsymbol{0}&\boldsymbol{0}&\boldsymbol{0}&\boldsymbol{0}&\boldsymbol{0}&\cdots&\boldsymbol{I}_2&\boldsymbol{0}
	\end{pmatrix}\begin{pmatrix}
		\boldsymbol{x}_t^{(1)}\\ \boldsymbol{v}_t^{(1)}\\ \boldsymbol{x}_t^{(2)}\\ \boldsymbol{v}_t^{(2)}\\ \vdots\\ \boldsymbol{x}_t^{(23)}\\ \boldsymbol{v}_t^{(23)}\end{pmatrix}+\begin{pmatrix}
		\boldsymbol{\varepsilon}_t^{(1)}\\ \boldsymbol{\varepsilon}_t^{(2)}\\ \vdots \\ \boldsymbol{\varepsilon}_t^{(23)}\end{pmatrix}.
\end{align*}
Here the measurement error vector is $(\boldsymbol{\varepsilon}_t^{(1)}\quad\boldsymbol{\varepsilon}_t^{(2)}\quad\cdots\quad\boldsymbol{\varepsilon}_t^{(23)})\sim\mathrm{N}(\boldsymbol{0},\boldsymbol{\Sigma})$ with $\boldsymbol{\Sigma}=\mathrm{Diag}(\sigma_{x,1}^2,\sigma_{y,1}^2,\sigma_{x,2}^2,\sigma_{y,2}^2,\dotsc,\sigma_{x,23}^2,\sigma_{y,23}^2)$ and the acceleration vector $(\boldsymbol{a}_t^{(1)}\cdots\boldsymbol{a}_t^{(23)})\sim\mathrm{N}(\boldsymbol{0},\boldsymbol{Q})$. 

It would be interesting to use this framework to model the interactions between different football players and the ball through the covariance matrix $\boldsymbol{Q}$; obviously, in a real match one expects a strong correlation between all entities. An unstructured~$ \boldsymbol{Q}$ consists of $46^2=2116$ parameters and adding the diagonal elements of $\boldsymbol{\Sigma}$ yields a total of $2162$ parameters. We found that this general case takes a prohibitively long time to optimize, and we have to simplify the problem by imposing additional structure on~$\boldsymbol{Q}$. To keep computations manageable, we disregard correlations between entities, by assuming that $\boldsymbol{Q}$ is a block diagonal matrix given by $\boldsymbol{Q}=\mathrm{BlockDiag}(\boldsymbol{Q}_1,\dotsc,\boldsymbol{Q}_{23})$ where $\boldsymbol{Q}_k=\mathrm{Var}(\boldsymbol{a}_t^{(k)})$ for $k=1,\dotsc,23$. In other words, each player's movement is modeled using his/her own state space equations that are independent of the other players.

\begin{figure}[ht!]
	\centerline{\includegraphics[scale=0.45]{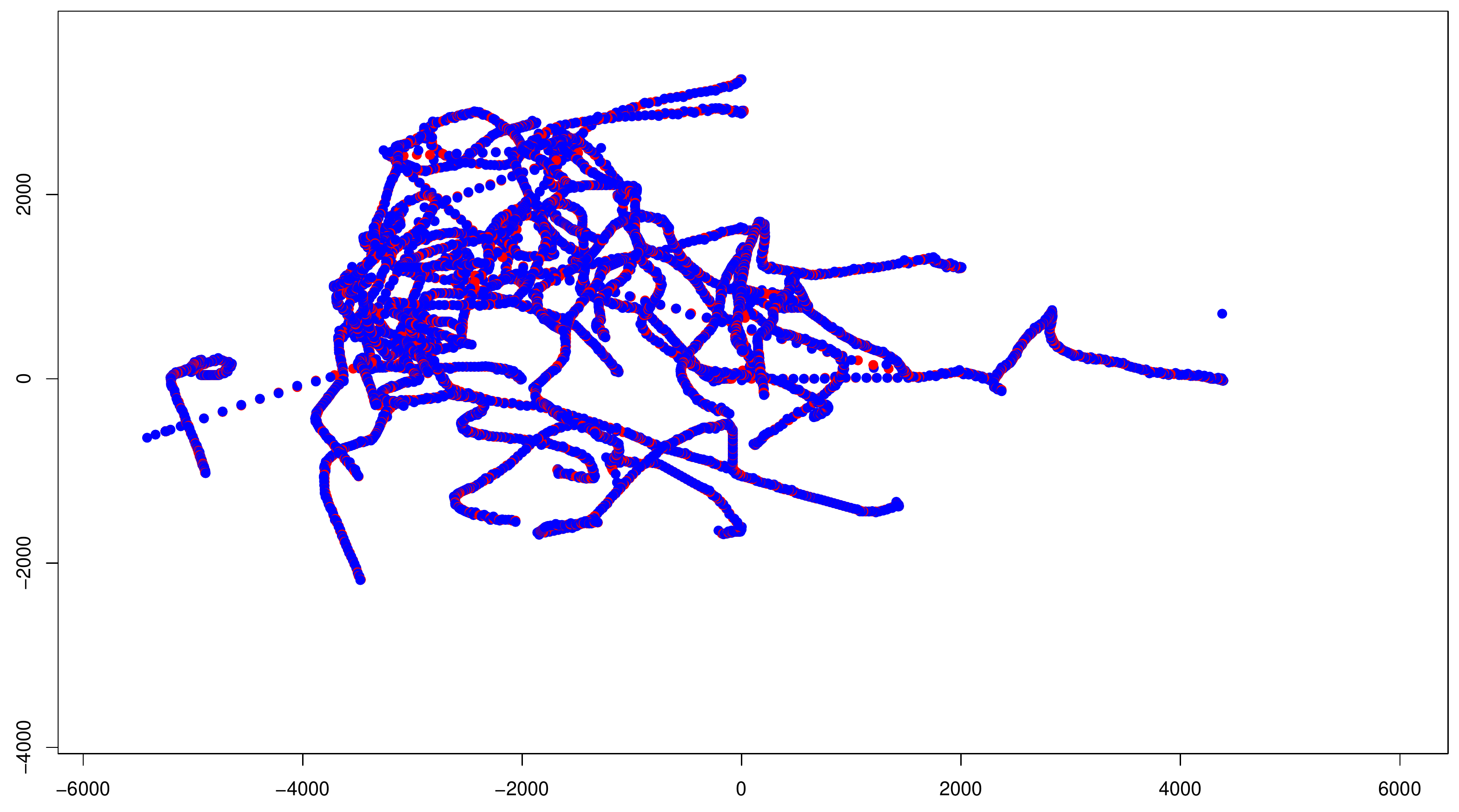}}
	\caption{One-step-ahead predicted positions for the ball and all $22$ players (blue) with their true paths (red). The path of the ball is the zig-zag dotted line.}
	\label{sci-fig:footballgame}
\end{figure}

If the prediction horizon is short, e.g., one step ahead, we found that this choice of $\boldsymbol{Q}$ gives reasonable predictive performance as shown in Figure \ref{sci-fig:footballgame}. Here we have used $5$ past time points to predict one timestep ahead and we see that the one-step-ahead predicted player's position (blue) closely follows the truth (red) over the span of $206$ time points. Moreover, the path of the ball is instantly recognizable as the zig-zag dotted line (due to it being the fastest object) embedded among the network of trajectories. If longer prediction horizons are sought, then this simplifying assumption might not give good performance and cross-covariance terms between players and ball are needed. To that end, one can consider low-rank approximations or imposing sparsity constraints on $\boldsymbol{Q}$. Alternatively, we can turn to machine-learning methods by training a (deep) multi-level neural network to learn these complex interactions; this is the subject of the next section.

\section{Methods: data-driven}
\label{sci-sec:data-driven}

In this section we describe machine-learning techniques to model spatio-temporal trajectories of players and the ball throughout the game, in order to acquire meaningful insight on football kinematics.  Our philosophy is that we aim to construct networks that can \emph{generate} trajectories that are \emph{statistically indistinguishable} from the actual data.  Successfully trained networks of this type have a number of benefits. They allow one to quickly generate more data;  the components of such networks can be re-used (we show an example in Section~\ref{sci-sec:dis}); when they produce `latent spaces', then these latent spaces may be interpreted by humans; and the structure of succesful networks and the values of the trained parameters should, in theory, give information about the trajectories themselves.

In Section \ref{sci-sec:gan}, we use Generative Adversarial Networks, such that two networks are pitted against each other to generate trajectories. Next, in Section \ref{sci-sec:vae}, we consider another class of networks called Variational Autoencoders, where we do data compression and train the network to replicate trajectories by learning important features. Finally, in Section~\ref{sci-sec:dis} we investigate a method to discriminate between walking patterns of two different football players.

\subsection{Generative Adversarial Network}\label{sci-sec:gan}
Generative Adversarial Networks (GANs) are deep neural net architectures introduced by \cite{sci-bib:Goodfellow2014} which exploit the competition between two (adversarial) networks: a generative network called the Generator and a discriminative network called the Discriminator.

Both the Generator and Discriminator are trained with a training set of real observations, and against each other. The Discriminator is a classifier; it has to learn to differentiate between real and generated observations, labeling them as ``realistic'' and ``fake'' respectively. The Generator, on the other hand, has to learn to reproduce features of the real data and generate new observations which are good enough to fool the Discriminator into labeling them as ``realistic''.

\subsection*{2D positional data into images}

GANs have been used with great success in image recognition, 3D-models reconstruction and photorealistic imaging; see e.g. \cite{sci-bib:karazeev}. Because of the limited time available to us, we decided to capitalize on existing codes for images; we use~\cite{sci-bib:oreilly}.
By rescaling the data accordingly we map the football field to the square $[-1,1]^2$ and interpret a $10$ seconds trajectory as a $2\times100$ gray-scale image: for each of the 100 time points, the two degrees of freedom indicate the rescaled $x$- and $y$-positions. This ``image" is what we input into the neural network machinery. 

\subsection*{Network setup}
The algorithm we use is a repurposed version of the basic convolutional neural network found at \cite{sci-bib:oreilly}, which is meant to recognize and reproduce handwritten digits. There is a structural difference between the two:
\begin{itemize}
	\item the original algorithm works with the MNIST digit dataset, which consists of $28\times28$ black-and-white images of $10$ possible states (the digits $0$-$9$);
	\item our algorithm works with $2\times100$ gray-scale images, containing an aggregation of $10$ seconds of play.
\end{itemize}
If we were to convert our gray-scale images to black-and-white, we would lose too much information. 

\begin{figure}[ht]
	\centering
	\includegraphics[width=0.8\textwidth]{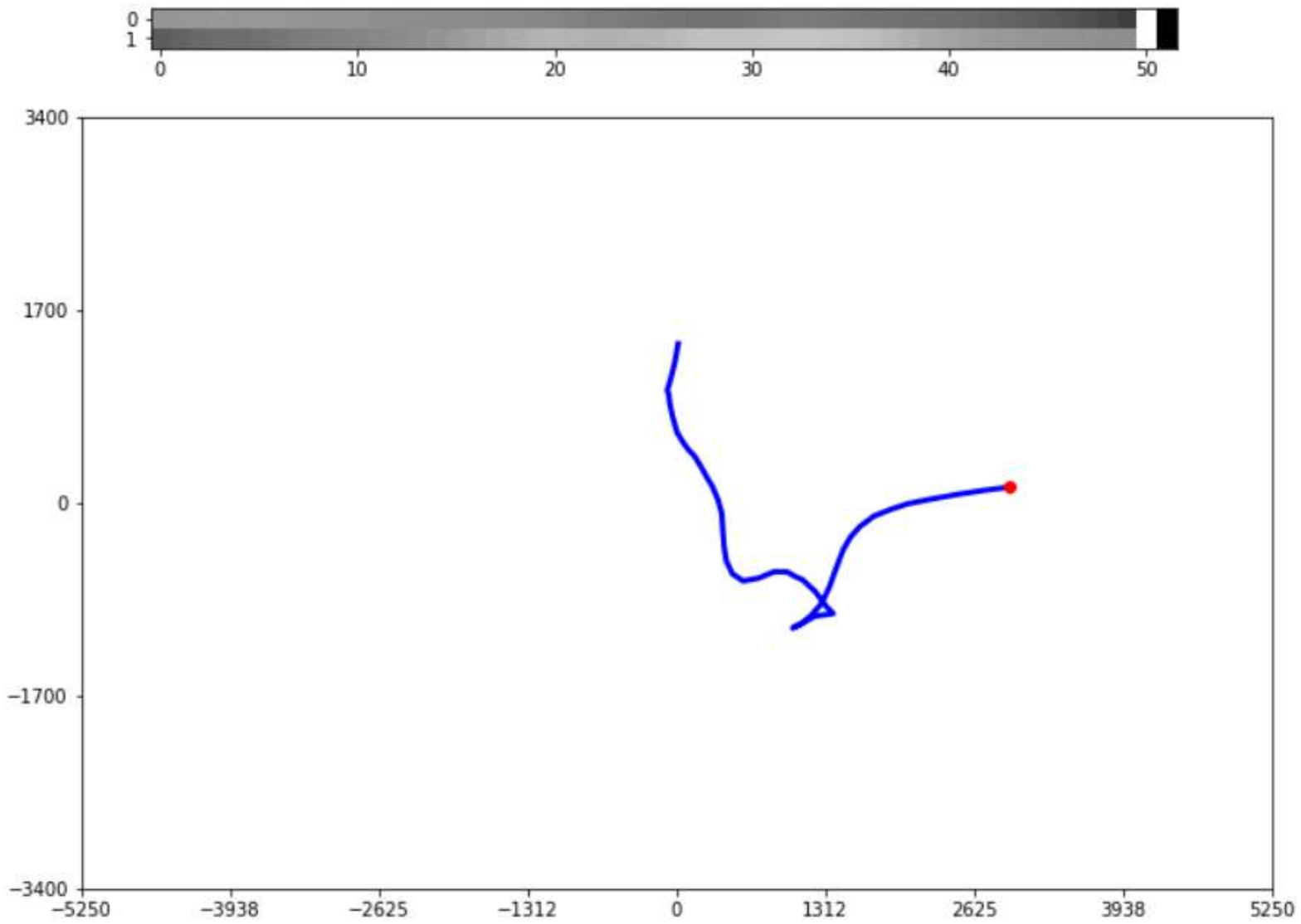}
	\caption{A non-trivial real trajectory and its twofold representation. The $(x,y)$-coordinates as gray-scale image (top) and the real trajectory on the football field (bottom).}
	\label{sci-fig:Real_example}
\end{figure}

Another important difference is in the intrinsic asymmetry of the data:
\begin{itemize}
	\item in the original version, both the Discriminator and the Generator look at $3\times3$ or $5\times5$ spatial features of the images: useful information about the topology of the shape can be obtained by looking at spatial neighborhoods of any given pixel;
	\item in our case we want to look a the $x$ and $y$ coordinates independently, therefore our Discriminator and Generator work with 
	one-dimensional temporal features: the information regarding the $x$- or $y$-trajectory in a temporal neighborhood of each position, i.e., its recent past and future. The information about the recent past and future of the trajectory should not be too small, otherwise the feature only observes the position of a player. On the other hand, if the feature is too large, it observes almost the entire 10-second trajectory, and the trajectory only contains a few features. To balance this trade-off we use $1\times5$ and $1\times10$ temporal features.
\end{itemize}
By making this tweak to the original algorithm we exploit the natural directionality of the data and we avoid overlapping the spatial properties (i.e., the shade of gray) and the temporal properties (i.e., the variation in shade). To have a sense of what this means we visualize the correspondence between the $(x,y)$-coordinates and the real trajectory of a player, see Figure \ref{sci-fig:Real_example}.

\subsection*{The algorithm}
We limit our training set to all random samplings of 20-second trajectories of any single player (excluding goalkeepers and the ball) during a single fixed match. This should give some extra structure for the network to work with while maintaining a diverse enough data sample.

\begin{figure}[ht]
	\centering
	\includegraphics[width=0.8\textwidth]{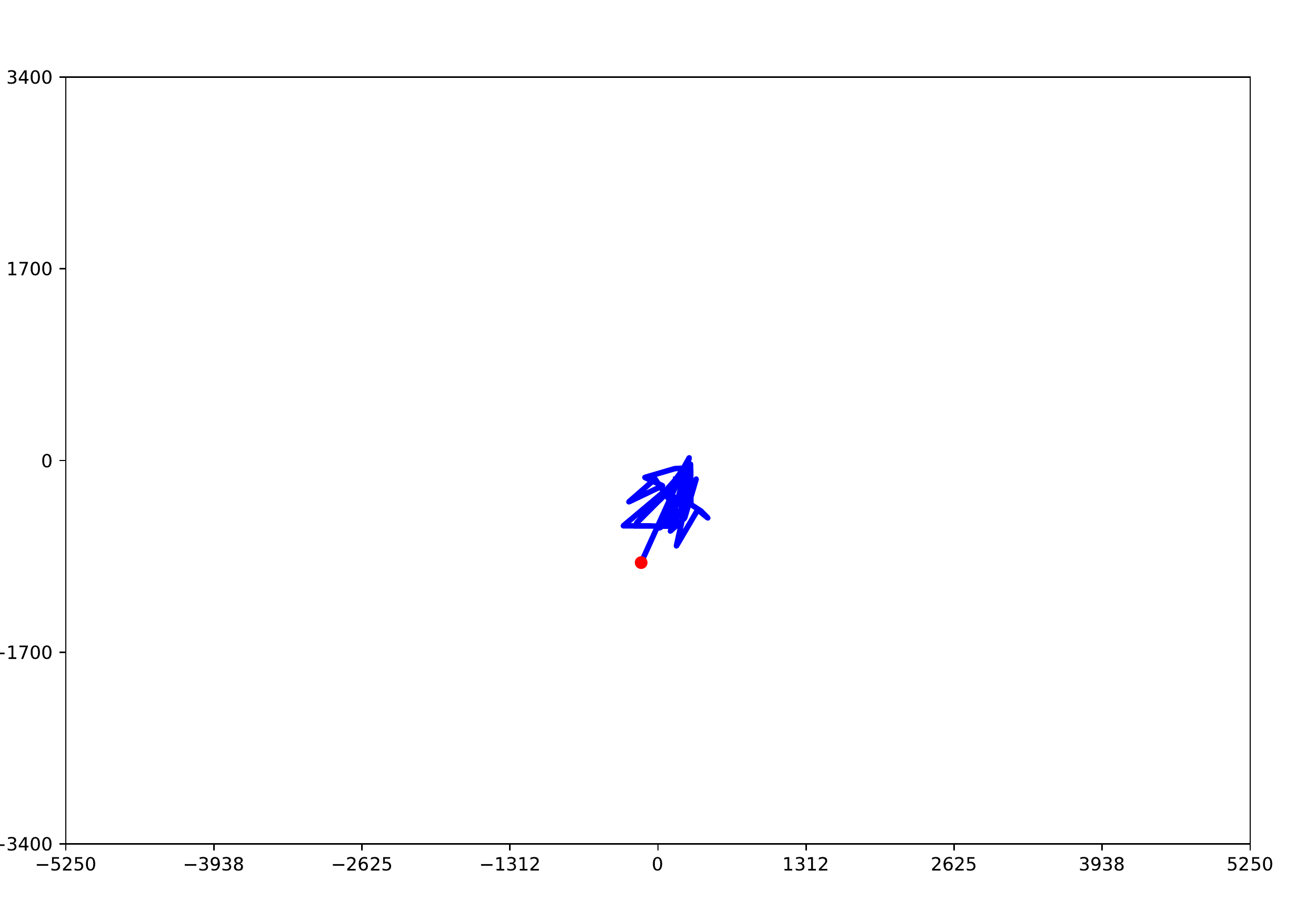}
	\caption{A trajectory from the untrained Generator.}
	\label{sci-fig:Untrained_Gen}
\end{figure}

The initialization of the parameters is the same as in the original algorithm, the Generator takes a standard Gaussian noise vector as input and then produces a new image based on the updates made by the network. To have a glance of what an untrained Generator is capable of, see Figure \ref{sci-fig:Untrained_Gen}.

The Discriminator is then pre-trained with real and generated trajectories. After this first training epoch, the Discriminator is able to correctly discriminate between the real trajectories and the untrained noisy ones produced by the Generator. Here an epoch consists of one full learning cycle on the training set. Then the main training session begins. From the second epoch and above, the Discriminator is trained with real and generated data and the Generator itself is trained against the Discriminator. This produces a Generator-Discriminator feedback loop that forces both networks to improve themselves with the objective to outperform the other. This is achieved by implementing a loss function to measure three quantities:
\begin{itemize}
	\item Discriminator loss vs real: it measures how far the Discriminator is from labeling a real trajectory as ``realistic'';
	\item Discriminator loss vs Generator: it measures how far the Discriminator is from labeling a generated image as ``fake'';
	\item Generator loss vs Discriminator: it measures how far the Discriminator is from labeling a generated image as ``realistic''.
\end{itemize}
The first loss function deals with the interaction between the Discriminator and the real world, it makes sure that the network is adapting to recognize new real observations. The second and third loss functions on the other hand, work against each other: one is trying to force the Discriminator to always label ``fake'' when presented with a generated image, while the other is forcing the Generator to produce data that mimics the Discriminator's perception of the real world. The loss function used throughout the algorithm is the cross-entropy loss, for a discussion see \cite{sci-bib:takeshi}.

\subsection*{Performance and limitations}
\begin{figure}[ht]
	\centering
	\begin{subfigure}[b]{.30\linewidth}
		\includegraphics[width=\linewidth]{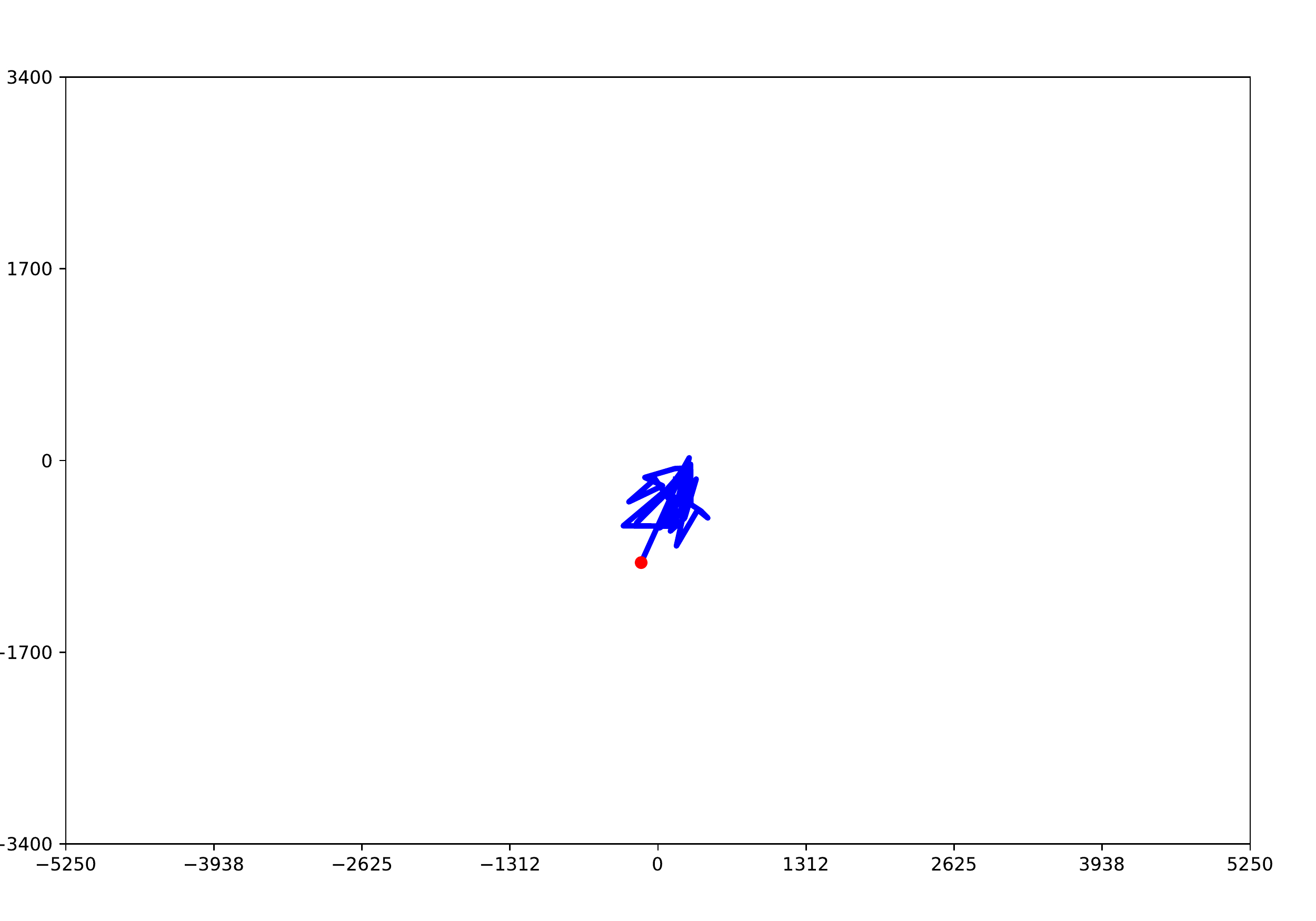}
	\end{subfigure}
	\begin{subfigure}[b]{.30\linewidth}
		\includegraphics[width=\linewidth]{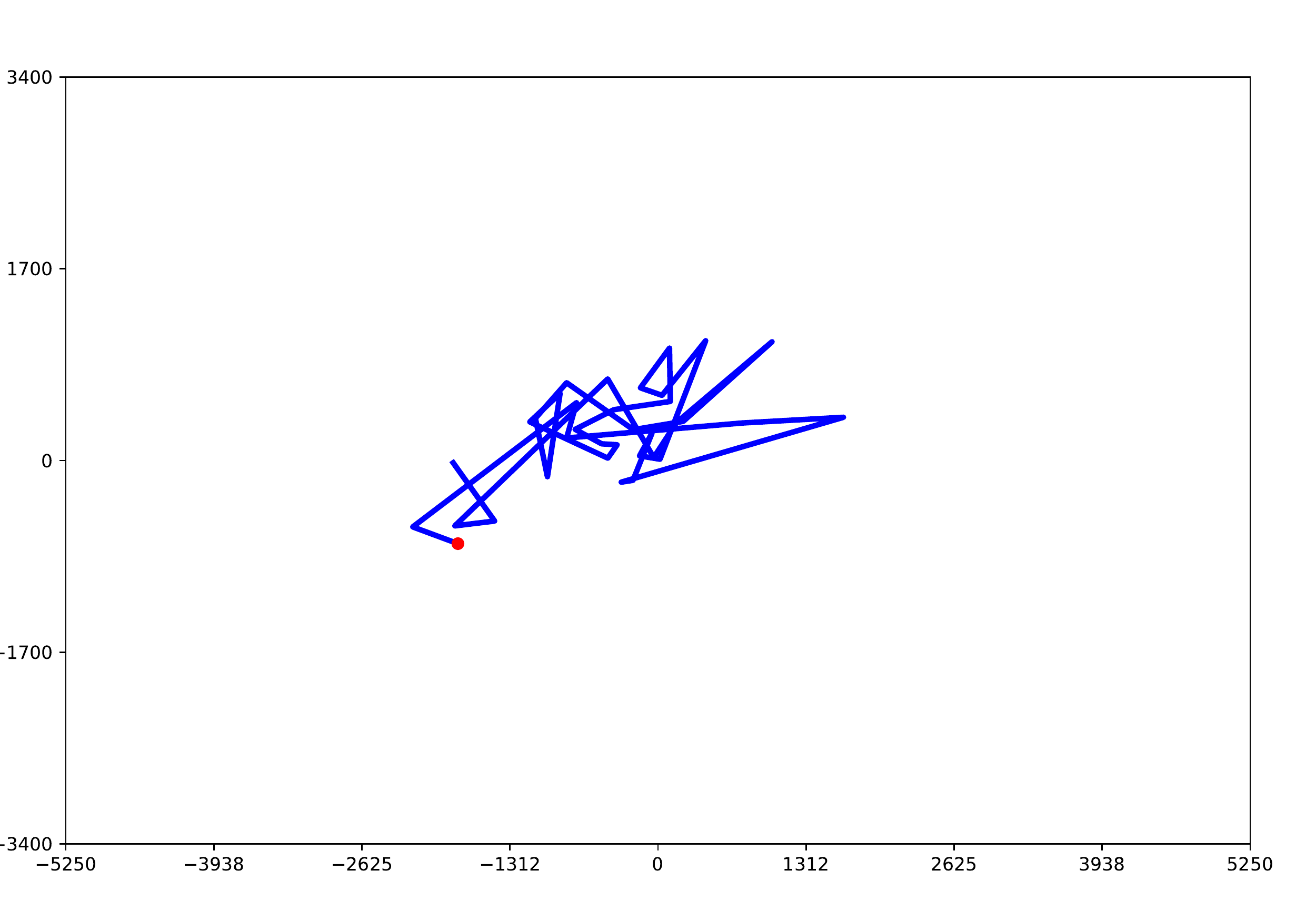}
	\end{subfigure}
	\begin{subfigure}[b]{.30\linewidth}
		\includegraphics[width=\linewidth]{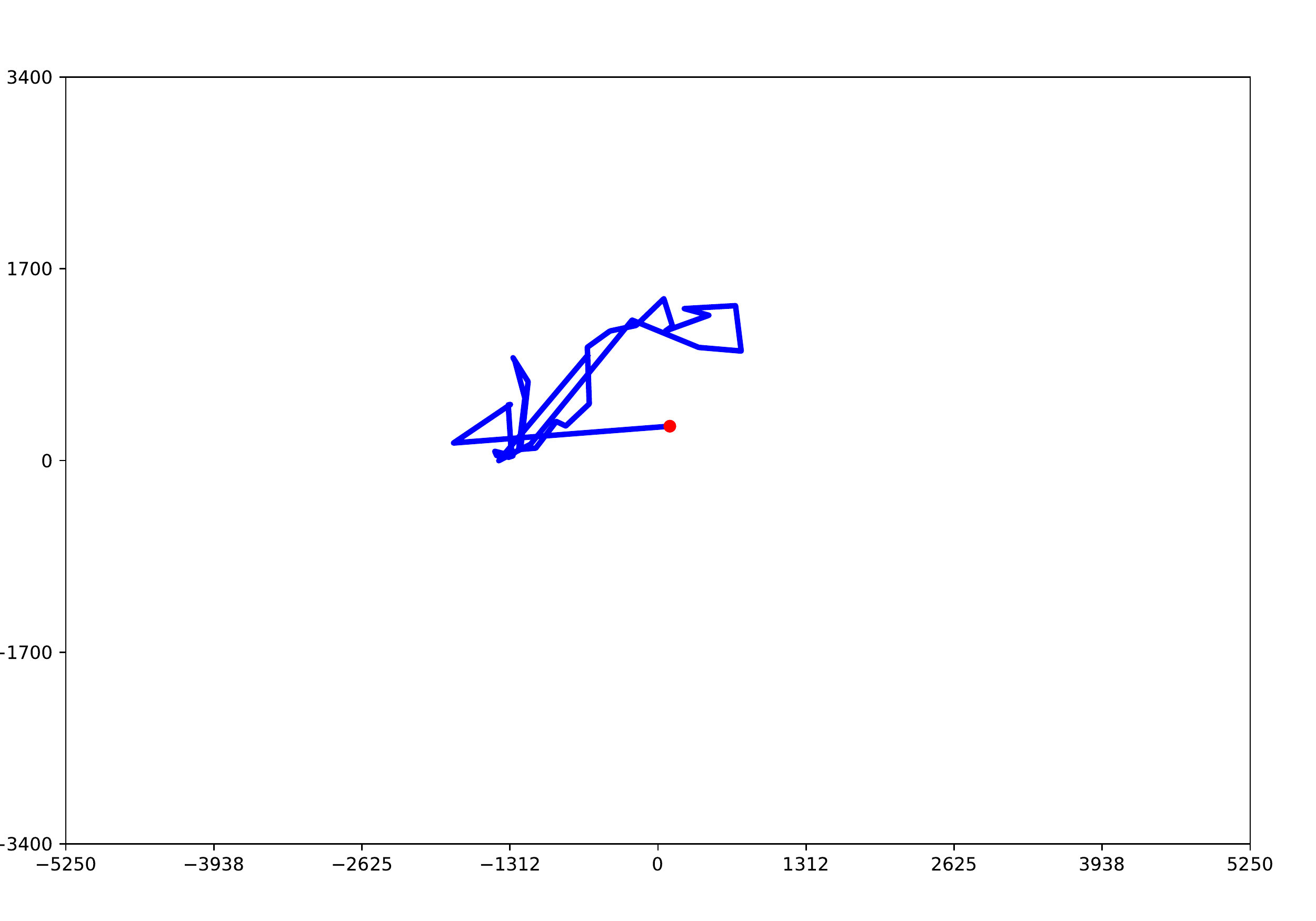}
	\end{subfigure}
	
	\begin{subfigure}[b]{.30\linewidth}
		\includegraphics[width=\linewidth]{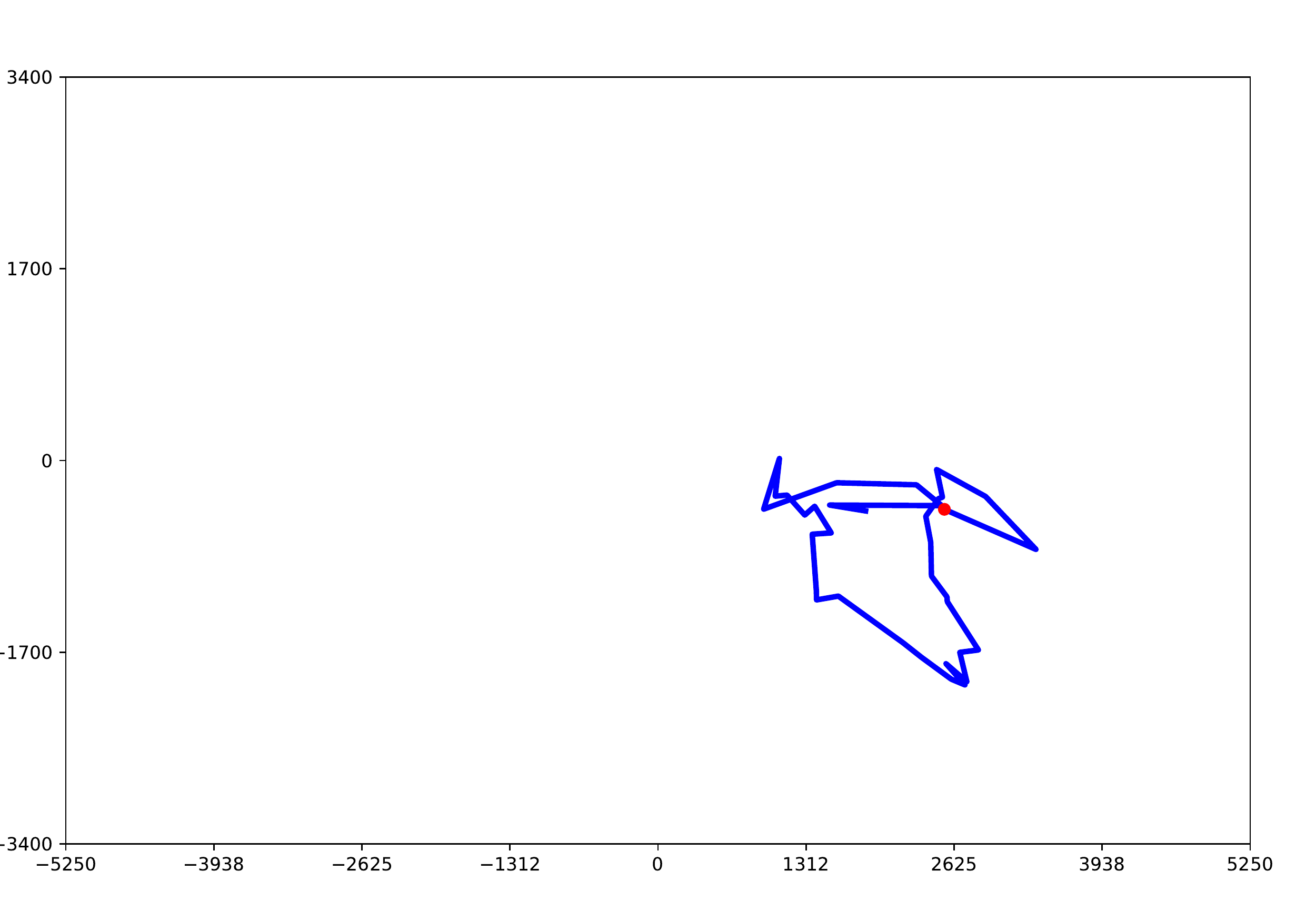}
	\end{subfigure}
	\begin{subfigure}[b]{.30\linewidth}
		\includegraphics[width=\linewidth]{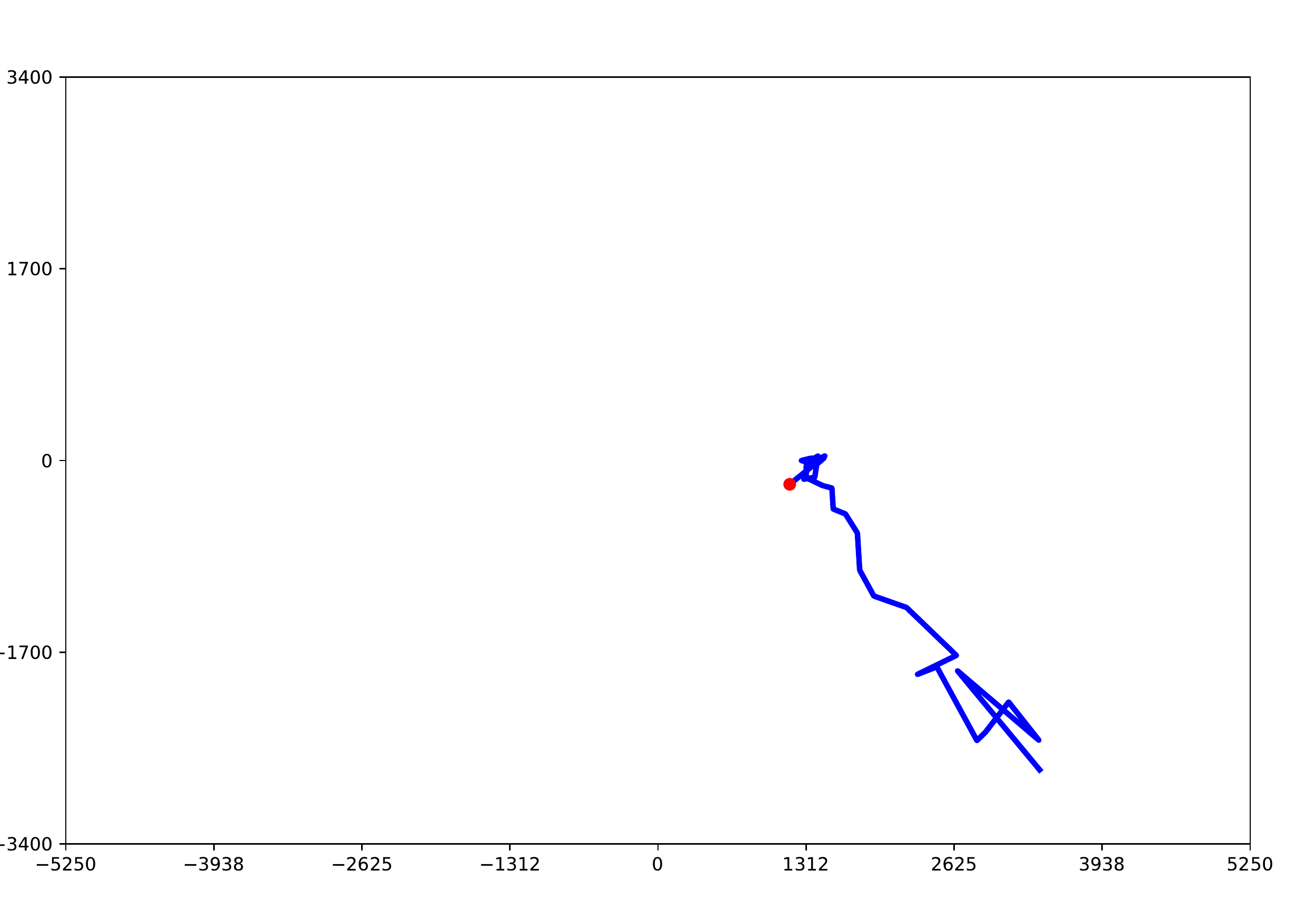}
	\end{subfigure}
	\begin{subfigure}[b]{.30\linewidth}
		\includegraphics[width=\linewidth]{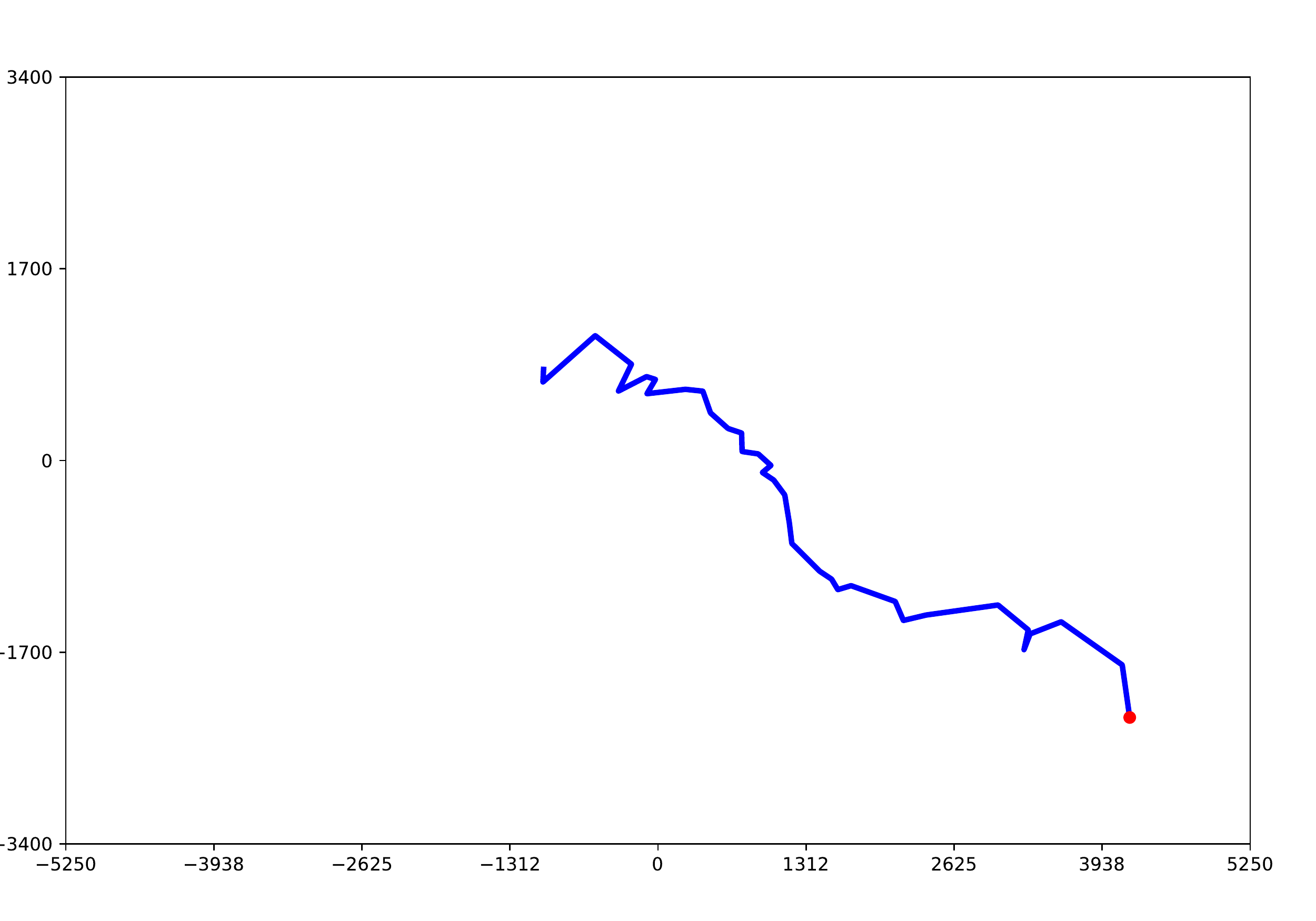}
	\end{subfigure}
	\caption{Different stages of GAN training (from left to right and from top to bottom). The network goes from random noise to shape recovery, but it is not able to filter out local noise consistently.}
	\label{sci-fig:Training}
\end{figure}

Properly training a GAN requires a long time and much can go wrong in the process. The Generator and Discriminator need to maintain a perfect balance, otherwise one will outperform the other causing either the Discriminator to blindly reject any generated image, or the Generator to exploit blind spots the Discriminator may have. After a training session of $15$ hours our GAN managed to go from random noise trajectories to smooth and structured ones, although not fully learning the underlying structure of the data. While the generated movements look impressive when compared to the untrained ones, they are still underperforming when confronted with the real world. First and foremost, the acceleration pattern of the players make no physical sense, i.e., the algorithm is not able to filter out local small noise, and the trajectories are not smooth enough. The evolution of the network during training is shown in Figure \ref{sci-fig:Training}. In the end the GAN is not consistent enough when asked to generate large samples of data: too many trajectories do not look realistic.

\subsection{Variational Autoencoder}\label{sci-sec:vae}
In parallel, we implemented a Variational Autoencoder (VAE) as introduced by \cite{sci-bib:vae}. Like a GAN, a VAE is an unsupervised machine-learning algorithm that gives rise to a generative model. 

We will apply the VAE algorithm on normalized trajectory data spanning 50 seconds. We call the set of all such trajectory data $X$. As the trajectories are sampled at intervals of 0.1 seconds, this means that we can identify $X$ with $[0,1]^{1000}$. 

A VAE consists of two neural networks, an encoder and a decoder. The encoder is a function (parametrized by a vector $\phi$)
\[
\mathsf{Enc}_\phi: X \times \mathcal{E} \to Z
\]
that maps from the product of the space $X$ of input data and a space of noise variables $\mathcal{E}$, to the so-called latent space $Z$. We identify the space $Z$ with $\mathbb{R}^d$ ($d=10$). 
The decoder is a function (parametrized by a vector $\theta$)
\[
\mathsf{Dec}_\theta : Z \times \Omega \to X
\]
which maps from the latent space $Z$ and a second space of noise variables $\Omega$ back to the data space $X$. 

We choose the spaces of noise variables $\mathcal{E}$ and $\Omega$ to be Euclidean, with the same dimension as $Z$ and $X$ respectively, and endow them with standard Gaussian measures. 

The encoder and decoder have a special structure. We implemented (as neural networks) functions
\[
\mu_{Z, \phi} : X \to Z \quad \text{ and } \quad \sigma_{Z, \phi} : X \to Z
\]
and chose
\[
\mathsf{Enc}_\phi(x,\epsilon) := \mu_{Z, \phi}(x) + \mathrm{diag}(\sigma_{Z, \phi}(x)) \epsilon.
\]
Here, $\mathrm{diag}(\sigma_{Z,\phi}(x))$ is a diagonal matrix with $\sigma_{Z, \phi}(x)$ on the diagonal. Equivalently, $\mathrm{diag}(\sigma_{Z, \phi}(x)) \epsilon$ is just the elementwise product of $\sigma_{Z,\phi}(x)$ and $\epsilon$. 

Similarly, we implemented a function 
\[
\mu_{X, \theta} : Z \to X
\]
and selected a constant $\sigma_X \in (0,\infty)$ and chose
\[
\mathsf{Dec}_\theta(z, \omega) := \mu_{X, \theta}(z) + \sigma_X \omega.
\]
The decoder provides us with a generative model for the data: to generate a data point we first sample $z$ and $\omega$ independently according to standard normal distributions, after which we apply the decoder to the pair $(z,\omega)$. Alternatively, we can generate zero-noise samples by only sampling $z$ and computing $\mathsf{Dec}_\theta(z,0)$.

The Variational Autoencoder $\mathsf{VAE}_{\phi,\theta}:X \times \mathcal{E} \times \Omega \to X$ is the composition of the encoder and decoder in the sense that
\[
\mathsf{VAE}_{\phi, \theta}(x,\epsilon,\omega) = \mathsf{Dec}_\phi ( \mathsf{Enc}_\theta(x, \epsilon) , \omega).
\] 
The parameters $\phi$ and $\theta$ of the VAE are optimized simultaneously, so that when we apply the VAE to a randomly selected triple of trajectory $x$, noise variable $\epsilon$ and noise variable $\omega$, the result is close to the original trajectory, at least on average. 

To this end, we follow \cite{sci-bib:vae} and minimize an average loss, for the loss function $\mathcal{L}_{\phi,\theta}:X \times \mathcal{E} \to \mathbb{R}$ given by
\begin{align}\label{sci-eq:loss}
	\frac{1}{\sigma_X^2} \mathcal{L}_{\phi, \theta}(x, \epsilon) 
	&:= \frac{1}{\sigma_X^2} \left\| x - \mu_{X, \theta}\big(\mathsf{Enc}_\phi(x,\epsilon) \big)\right\|^2+ \|\mu_{Z,\phi}(x)\|^2   - d \nonumber\\
	&\qquad 
	- \mathrm{tr} \big(\log ( \mathrm{diag}(\sigma_{Z,\phi}(x))^2\big)+  \mathrm{tr}\big(\mathrm{diag}(\sigma_{Z,\phi}(x))^2\big).
\end{align}
For a derivation of this loss function, we refer the reader to the Appendix.

We implemented the Autoencoder in the Keras library for Python (\citealp{sci-bib:keras}). The library comes with an example VAE which we took as a starting point. We introduced a hidden layer $H_E$ in the encoder and $H_D$ in the decoder, which we both identified with $\mathbb{R}^{400}$, and implemented the functions $\mu_{Z,\phi}$ and $\sigma_{Z,\phi}$ as
\[
\begin{split}
\mu_{Z,\phi} &= m_{Z,\phi} \circ h_{E,\phi}\\
\sigma_{Z,\phi} &= \exp \circ \, {l}_{Z,\phi} \circ h_{E,\phi}
\end{split}
\]
where $h_{E,\phi}: X \to H_E$ is the composition of an affine map and ReLu activation functions, the functions $m_{Z,\phi}, {l}_{Z,\phi} : H_E \to Z$ are linear and $\exp: Z \to Z$ is the exponential function applied componentwise. 

Similarly, 
\[
\mu_{X, \theta} = m_{X, \theta} \circ h_{D, \theta}
\]
where the function $h_{D,\theta}: Z \to H_D$ is again a composition of an affine map and ReLu activation functions and the function $m_{X,\theta}: H_D \to X$ is a composition of an affine map and sigmoid activation functions.

We trained the model, i.e.~we adjusted the parameters $\phi$ and $\theta$ to minimize the average loss, using the `rmsprop' optimizer in its default settings. Whether the model trained successfully or not did seem to depend crucially on the version of the libraries used. For the results presented below, we used Keras version 2.1.3 on top of Theano version 1.0.1. We first set $\sigma_X \approx 0.15$. After training for 1000 epochs, the average loss was slightly below $2$. 

We used the VAE to approximate trajectories. We sampled at random trajectories $x_i$ from the data, and compared them to their approximations 
\[
\hat{x}_i := \mathsf{VAE}_{\phi, \theta}(x_i, 0, 0).
\]
The average absolute deviation per coordinate per time-step (expressed as a ratio with respect to the dimensions of the playing field) was approximately $0.02$, the average squared error per coordinate per time step was approximately $0.0008$ and the average maximum error per coordinate, taken over the whole trajectory, was less than $0.09$. 

\begin{figure}[h]
	\centering
	\includegraphics{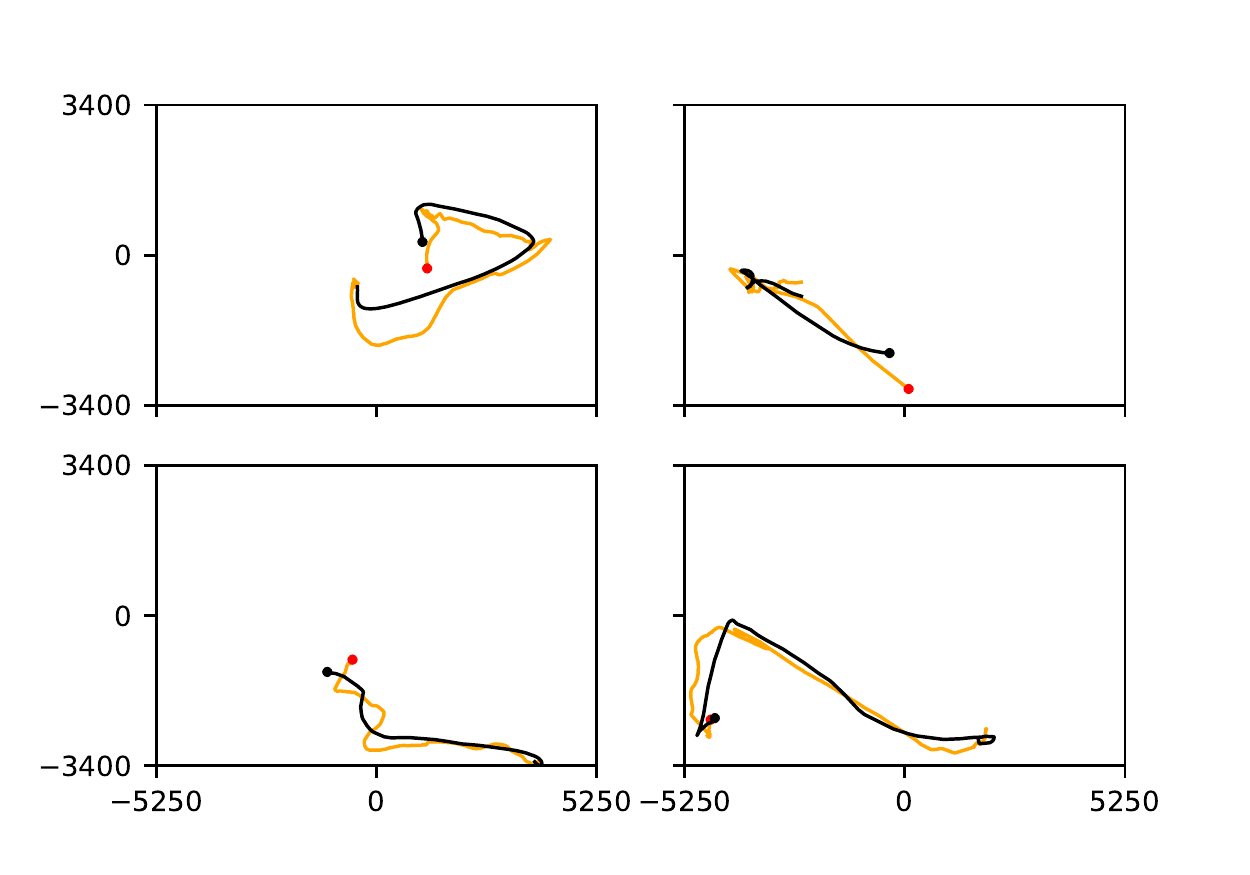}
	\caption{A collection of sampled trajectories (orange) and an approximation calculated by the VAE (black). In general, the approximating trajectories are much smoother. We chose $\sigma_X \approx 0.15$ in training the VAE.}
	\label{sci-fig:traj-reproduction}
\end{figure}

In Figure \ref{sci-fig:traj-reproduction} we show the result of sampling four random trajectories $x_i$ from the data, and comparing them to their approximation by the VAE. The approximating trajectories are much smoother than the original ones. 
Some qualitative features of the original paths, such as turns and loops, are also present in the approximating paths. Even though the average error in the distance per coordinate per time step is relatively small, visually there is still quite some deviation between the true and the approximating trajectories. We expect, however, that with a more extensive network, consisting of more convolutional layers, we can greatly improve the approximation.

\begin{figure}[h]
	\centering
	\includegraphics{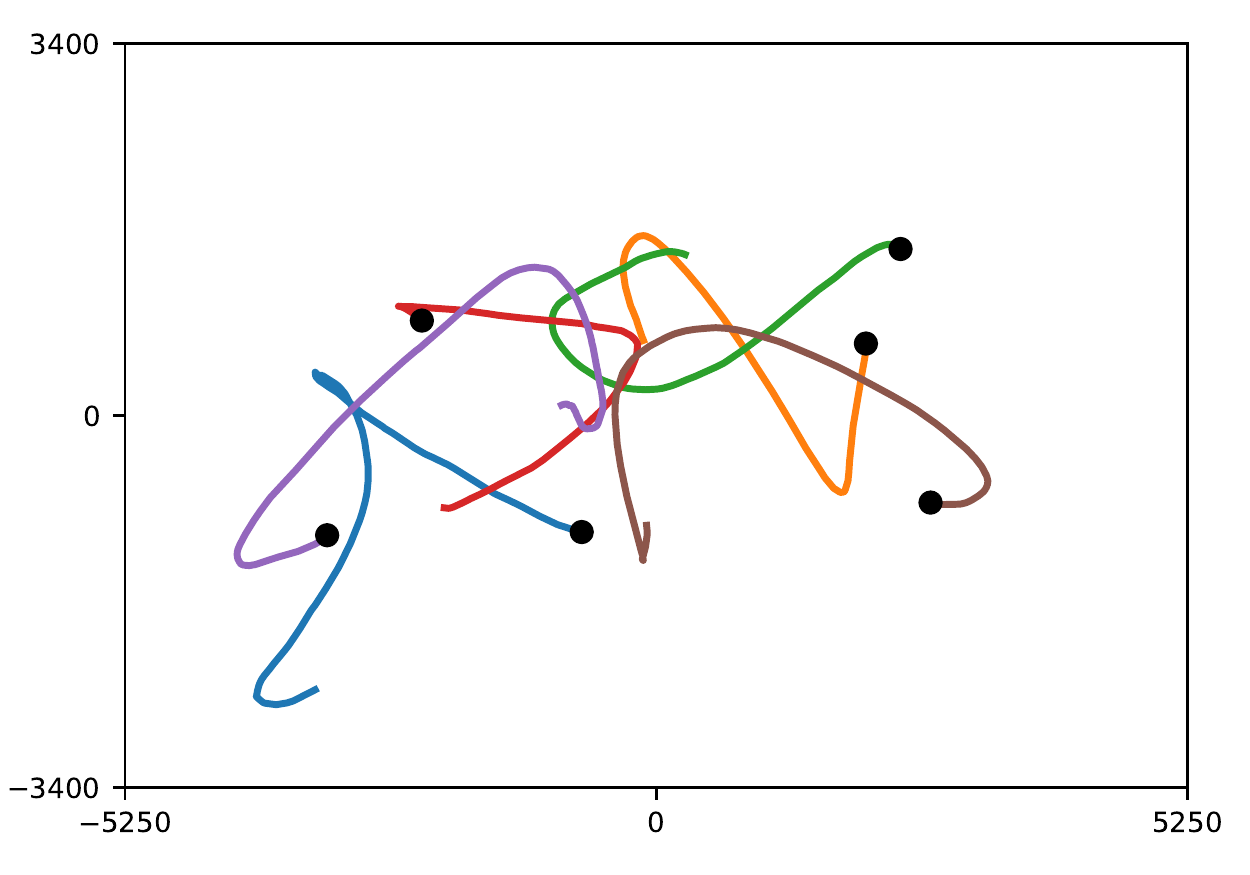}
	\caption{Six random trajectories generated by the generative model, i.e.~by the decoder part of the VAE.}
	\label{sci-fig:traj-generation}
\end{figure}

Next, we use the decoder of the VAE as a generative model. In particular, we sample trajectories in $X$ at random by first sampling $z\in Z$ according to a standard normal distributions, and computing the trajectory $\mathsf{Dec}_\theta(z, 0)$. A collection of six trajectories generated in this way is shown in Figure \ref{sci-fig:traj-generation}. At first sight, the generated trajectories look like they could have been real trajectories of football players. However, they are in general smoother than the real trajectories. We could also have generated trajectories by sampling both $z$ and $\omega$ according to standard normal distributions and computing $\mathsf{Dec}_\theta(z,\omega)$. However, those trajectories would have been much too noisy.

If we reduce the value of $\sigma_X$ to approximately $0.008$ and retrain the model, the approximation of the trajectories becomes slightly better, and the final average loss reduces to 0.67 after training for 600 epochs. The corresponding plots look similar to Figure \ref{sci-fig:traj-reproduction}. However, if we now use the decoder to generate trajectories, most of the trajectories end up close to the boundary of the playing field: the dynamics of the generated trajectories is then clearly very different from the original dynamics. 

In Appendix \ref{sci-sec:appendix}, we explain this effect by investigating the different parts of the loss function given in \eqref{sci-eq:loss}. The upshot is that when $\sigma_X$ is very small, the proportion of latent variables $z \in Z$ that are in the range of the encoder is very small (measured with the Gaussian measure on $Z$). If one applies the decoder to a $z \in Z$ which is in the range of the encoder, one probably gets a realistic trajectory. But for latent variables $z$ not in the range of the encoder, there is no reason for the decoded trajectories to look realistic at all. 

\subsection{Discriminator}\label{sci-sec:dis}
In the previous sections, we studied several methods to create generative models for the movement trajectories of football players, with the aim of capturing the underlying dynamics and statistics. In this section, we study to what extent movement trajectories of different soccer players can be distinguished. To this end,
we test the Discriminator network of the GAN introduced in Section~\ref{sci-sec:gan} on data of different soccer players. We train the Discriminator on the data of two soccer players, and then test if the Discriminator is able to distinguish their motion patterns. 
The success rate of the Discriminator to distinguish one player from the other then gives some insight in how different are the movement behaviors of two different players. 

The loss function for the Discriminator is the same as in Section~\ref{sci-sec:gan}. The data we use as input for the Discriminator are $(x,y)$-coordinates of 10-second player trajectories. We test the Discriminator on these unedited $(x,y)$-trajectories, and on centered $(x,y)$-trajectories, where the coordinates of each trajectory are centered such that the first coordinate always equals $(0,0)$. Thus, by using the uncentered data, the Discriminator may distinguish two players by using their positions on the field, whereas the Discriminator can only use movement patterns of particular players when the centered data are used. 

\begin{figure}[tb]
	\centering
	\begin{subfigure}{0.45\linewidth}
		\centering
		\includegraphics[width=\textwidth]{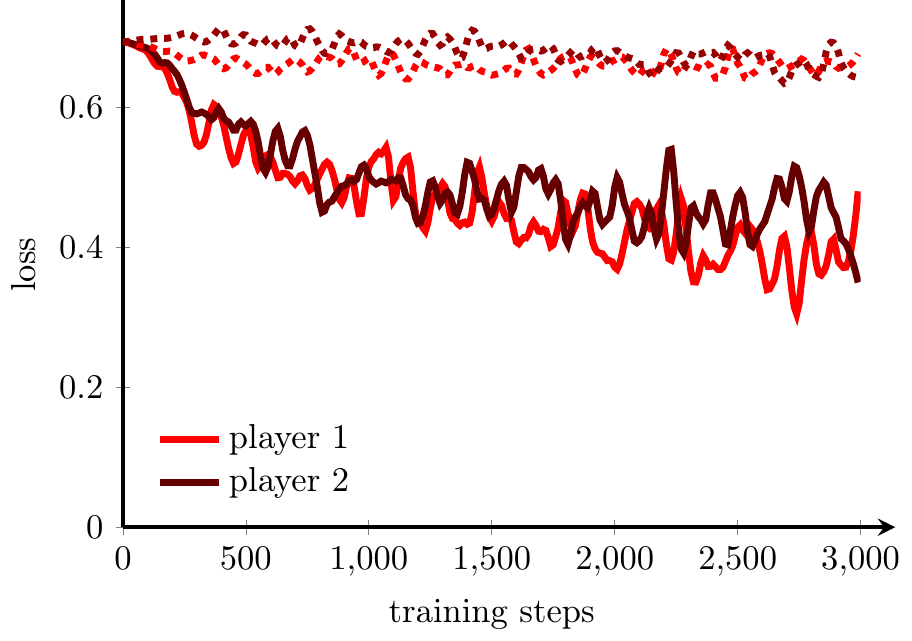}
		\caption{example 1}
		\label{sci-fig:ex1}
	\end{subfigure}
	\begin{subfigure}{0.45\linewidth}
		\centering
		\includegraphics[width=\textwidth]{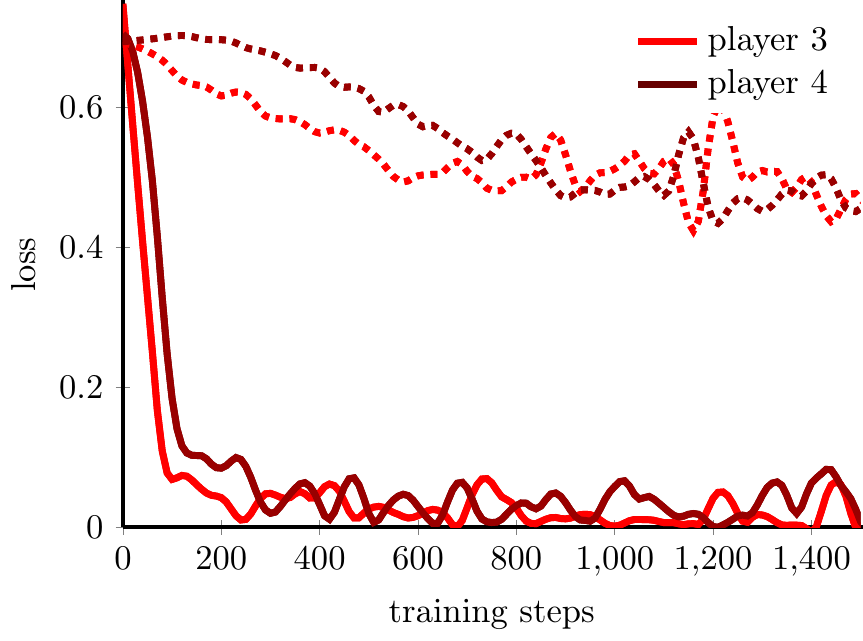}
		\caption{example 2}
		\label{sci-fig:ex2}
	\end{subfigure}
	\caption{Two examples of the Discriminator loss function for both players as a function of the number of training steps. The solid lines are the results for uncentered data and the dashed lines contain the results for the centered data. The two examples contain four different players.}
	\label{sci-fig:loss}
\end{figure}

Figure~\ref{sci-fig:loss} shows the Discriminator loss function for both players as a function of the number of training steps for two different sets of two players. We see that the loss function declines more for the uncentered data than for the centered data. Thus, the Discriminator distinguishes uncentered trajectories based on the location on the field where the movement pattern happens. The two different examples also show that it is easier to distinguish some players than others. 
Table~\ref{sci-tab:succrate} shows the success rate of correctly identifying the player corresponding to a given trajectory after the training period for the two sets of players of Figure~\ref{sci-fig:loss}. The success rate of the Discriminator using the uncentered data is higher than for the centered data in both examples. Using the centered data, the Discriminator has difficulties distinguishing between players 1 and 2 in the first example. In the second example, the success rate is much higher. Thus, some players display more similarities in their movement patterns than other players.

\begin{table}[htbp]
	\centering
	\begin{tabular}{rrrrrr}
		\toprule
		\textbf{} & \textbf{} & \textbf{Player 1} & \textbf{Player 2}& \textbf{Player 3} & \textbf{Player 4} \\
		\midrule
		\multicolumn{1}{c}{\multirow{2}[0]{*}{example 1}} & non-centered & 0.74  & 0.9 \\
		\multicolumn{1}{c}{} & centered & 0.2   & 0.96 \\
		\multicolumn{1}{c}{\multirow{2}[0]{*}{example 2}} & non-centered &&& 0.98     & 0.82 \\
		\multicolumn{1}{c}{} & centered &&& 0.54  & 0.95 \\
		\bottomrule
	\end{tabular}%
	\caption{The success rate of the Discriminator after training on the two examples of Figure~\ref{sci-fig:loss}. We use separate data sets for training and validation.}
	\label{sci-tab:succrate}%
\end{table}%

\section{Conclusion and future work}\label{sci-sec:con}
We used several methods to learn the spatio-temporal structure of trajectories of football players. With the state-space modeling approach we extracted velocity information from the trajectory data, and learned basic statistics on the motion of individual players. 
With deep generative models, in particular Variational Autoencoders, we captured the approximate statistics of trajectories by encoding them into a lower dimensional latent space. Due to limitations on time and computational power, we did not manage to successfully train Generative Adversarial Nets on the data. Nonetheless, we were able to use the Discrimator network to distinguish between different football players based on their trajectory data. The algorithm was more successful if we used non-centered rather than centered data, and was better at distinguishing between some players than others.

It is very likely that with deeper convolutional neural networks, we can train VAEs that approximate the statistics of the player trajectories even better. 
Besides, the approach can easily be extended to approximate trajectories of multiple players and the ball, although we may need more data to get an accurate model.

A big challenge is to interpret the latent space of the VAE. Ideally, one would be able to recognize qualities of the players as variables in the latent space. Although this is a difficult task in general, we expect that by adding additional structure in the architecture of the VAE, we can at least extract some relevant performance variables per player and recognize differences between players. Moreover, we could unify state-space models with VAEs to increase the interpretability of the latent variables.

By continuing this line of work, we could conceivably find an appropriate state space such that the football game can be fitted into a Reinforcement Learning framework. This framework may then be used to find optimal strategies, and to extract individual qualities of football players.  

\appendix

\section{Derivation of loss function of VAE}\label{sci-sec:appendix}

In this appendix we will derive the loss function for the Variational Autoencoder. The loss function is the same as the one used by \cite{sci-bib:vae}, and more generally corresponds to the usual loss function in variational inference, but our presentation here is slightly nonstandard and is based on general measure theoretic probability.

Before we can discuss the loss function and its meanings, we need to introduce notations for the various measures encountered in the problem. Both the encoder and the decoder of the VAE will induce measures on the product space $Z \times X$, and the optimization procedure will aim to bring these measures as close as possible to each other. We will first describe the encoder and the decoder measures.

\subsection*{Encoder measure}

Recall from Section \ref{sci-sec:vae} that we can identify $Z$ and $\mathcal{E}$ with $\mathbb{R}^d$. In addition, we let $X$ and $\Omega$ be subsets of $\mathbb{R}^k$ and we set $k=1000$ and $d=10$ in our own implementation. Let us start by assuming that trajectories are obtained by sampling independently according to a distribution $\mathbb{Q}_X$, which we assume to be absolutely continuous with respect to the $k$-fold product of Lebesgue measures $\mathcal{L}^k$ on $X$ with density $q_X:X \to [0,\infty)$. We denote the standard Gaussian measure on $\mathcal{E}$ by $\mathbb{Q}_\mathcal{E}$. The encoder induces a measure $\mathbb{Q}^\phi_{Z \times X \times \mathcal{E}}$ on the space $Z \times X \times \mathcal{E}$ by 
\[
\mathbb{Q}^\phi_{Z \times X \times \mathcal{E}} := (\mathsf{Enc}_\phi \times \mathrm{id} )_\# (\mathbb{Q}_X \otimes \mathbb{Q}_\mathcal{E})
\]
where $\mathrm{id}: X \times \mathcal{E} \to X \times \mathcal{E}$ is the identity map, and $g_{\#}Q$ is the pushforward measure of $Q$ induced by measurable function $g$ such that $(g_{\#}Q)(\mathcal{A})=Q(g^{-1}(\mathcal{A}))$ for any measurable set $\mathcal{A}$. Equivalently, for every bounded and continuous function $f: Z \times X \times \mathcal{E} \to \mathbb{R}$ it holds that
\[
\int_{Z \times X \times  \mathcal{E}} f d \mathbb{Q}^\phi_{Z \times X \times \mathcal{E}} = \int_{X \times \mathcal{E}} f(\mathsf{Enc}_\phi(x, \epsilon), x, \epsilon) d ( \mathbb{Q}_X \otimes \mathbb{Q}_\mathcal{E} )(x,\epsilon).
\]

We observe that $\mathbb{Q}_X$ and $\mathbb{Q}_\mathcal{E}$ are indeed the marginals of the measure $\mathbb{Q}^\phi_{Z \times X \times \mathcal{E}}$, and similarly we will denote by $\mathbb{Q}^\phi_{Z \times X}$ the $Z\times X$-marginal of $\mathbb{Q}^\phi_{Z \times X \times \mathcal{E}}$ etc.. We will occasionally refer to $\mathbb{Q}^\phi_{Z \times X}$ as the \emph{encoder measure} or the \emph{recognition model}. 

Finally, we denote the conditional distribution on $Z$ induced from the encoder given $x \in X$ by
\[
\mathbb{Q}^\phi_{Z|x} := \mathsf{Enc}_\phi(x, \cdot)_\# \mathbb{Q}_{\mathcal{E}}.
\]
We assume that its density with respect to $\mathbb{P}_{Z}$, the standard Gaussian measure on $Z$, exists and we denote it by $q^\phi_{Z|x}$. The measure $\mathbb{Q}^\phi_{Z\times X}$ is then absolutely continuous with respect to $\mathbb{P}_Z \otimes \mathcal{L}^k$ with density
\[
q^\phi_{Z \times X}(z,x) = q^\phi_{Z | x}(z) q_X(x).
\]
\subsection*{Decoder measure}

Analogously, we denote by $\mathbb{P}_Z$ and $\mathbb{P}_\Omega$ the standard Gaussian measures on $Z$ and $\Omega$ respectively. The decoder induces a measure $\mathbb{P}^\theta_{Z \times \Omega \times X}$ on the space $Z \times \Omega \times X$, given by
\[
\mathbb{P}^\theta_{Z \times \Omega \times X} := (\mathrm{id} \times \mathsf{Dec}_\theta)_\# (\mathbb{P}_Z \otimes \mathbb{P}_\Omega).
\]

Again, we observe that $\mathbb{P}_Z$ and $\mathbb{P}_\Omega$ are the marginals of $\mathbb{P}^\theta_{Z \times \Omega \times X}$ and we denote by $\mathbb{P}^\theta_{Z \times X}$ the marginal probability distribution on $Z \times X$. We refer to $\mathbb{P}^\theta_{Z \times X}$ as the \emph{decoder measure} or the \emph{generative model}.
We will assume that $\mathbb{P}^\theta_{Z \times X}$ is absolutely continuous with respect to the product measure $\mathbb{P}_Z \otimes \mathcal{L}^k$, and that its density $p^\theta_{Z \times X} : Z \times X \to (0,\infty)$ is strictly positive. Since $\mathbb{P}_Z$ is the marginal of $\mathbb{P}^\theta_{Z \times X}$ it follows that the marginal density $p_Z:Z \to [0,\infty)$ is defined by
\[
p_Z(z) := \int_X p^\theta_{Z\times X}(z,x) d \mathcal{L}^k(x)
\]
and $p_Z(z) = 1$ for every $z \in Z$. We also define the conditional density
\[
p^\theta_{X|z}(x) = \frac{p^\theta_{Z\times X}(z,x)}{p_Z(z)}.
\]
We denote the corresponding conditional probability distribution on $X$ by $\mathbb{P}^\theta_{X|z}$ and note that it coincides with the law of the decoder conditioned on $z \in Z$,
\begin{equation}
	\label{sci-eq:PXthetaz}
	\mathbb{P}^\theta_{X|z} = \mathsf{Dec}_{\theta}(z, \cdot)_{\#} \mathbb{P}_\Omega.
\end{equation}
In the particular context of the Variational Autoencoder explained in Section \ref{sci-sec:vae}, we find that
\[
p^\theta_{Z \times X}(z,x) = p^\theta_{X|z}(x)p_Z(z) =
\frac{1}{(2 \pi \sigma_X^2)^{k/2}}
\exp\left( - \frac{1}{2\sigma_X^2} \| x - \mu_{X,\theta}(z) \|^2\right).
\]

Similarly, we define the marginal density $p^\theta_X: X \to [0,\infty)$ by
\[
p^\theta_{X}(x) := \int_Z p^\theta_{Z \times X}(z,x) d \mathbb{P}_Z(z).
\]
Note that $p^\theta_{X}(x) > 0$ for all $x \in X$. 
We denote the associated probability distribution on $X$ by $\mathbb{P}^\theta_X$. We set
\[
p^\theta_{ Z |x } (z) := \frac{p^\theta_{Z\times X}(z,x)}{p^\theta_{X}(x)}
\]
and denote by $\mathbb{P}^\theta_{Z|x}$ the associated conditional probability distribution that has density $p^\theta_{Z|x}$ with respect to $\mathbb{P}_Z$. 

Note that by definition, the following version of Bayes' Theorem holds
\begin{equation}
	\label{sci-eq:BayesDefinition}
	p^\theta_{Z|x}(z) p^\theta_X(x) = p^\theta_{Z \times X}(z,x) = p^\theta_{X|z} (x) p_Z(z).
\end{equation}

\subsection*{Derivation of loss function}

The loss function of the Variational Autoencoder is built around the relative entropy, or more commonly known as the Kullback-Leibler (KL) divergence. If $\mathbb{P}$ and $\mathbb{Q}$ are probability measures on a measure space $Y$, the Kullback-Leibler divergence $D_{\mathrm{KL}}(\mathbb{Q} \| \mathbb{P})$ is defined to be $+\infty$ if $\mathbb{Q}$ is not absolutely continuous with respect to $\mathbb{P}$, and otherwise
\[
D_{\mathrm{KL}}(\mathbb{Q} \| \mathbb{P}) := \int_Y \log \frac{d \mathbb{Q}}{d \mathbb{P}} d \mathbb{Q},
\]
where $\tfrac{d \mathbb{Q}}{d \mathbb{P}}$ is the Radon-Nikodym derivative which we can take as the density of $\mathbb{Q}$ with respect to $\mathbb{P}$.

We aim to minimize over all $\theta$ and $\phi$ an approximation of
\[
D_{\mathrm{KL}}( \mathbb{Q}^\phi_{Z \times X} \| \mathbb{P}_{Z \times X}^\theta ).
\]
This has the interpretation that we search for $\theta$ and $\phi$ so that it is hard to distinguish the encoder distribution $\mathbb{Q}^\phi_{Z\times X}$ from the decoder distribution $\mathbb{P}^\theta_{Z \times X}$.

In view of Bayes' Theorem given by \eqref{sci-eq:BayesDefinition}, we can write this KL divergence in different ways as follows
\begin{align}
	\nonumber D_{\mathrm{KL}}(\mathbb{Q}^\phi_{Z \times X} \| \mathbb{P}^\theta_{Z \times X}) 
	&= \int_{Z \times X} \log \frac{q^\phi_{Z \times X}(z,x)}{p^\theta_{Z \times X}(z,x)} d \mathbb{Q}^\phi_{Z \times X}(z,x)\\
	\label{sci-eq:KL-after-bayes}&= \int_{Z \times X} \log \frac{q^\phi_{Z|x}(z) q_X(x)}{p^\theta_{X|z}(x) p_Z(z)} d \mathbb{Q}^\phi_{Z \times X}(z,x)\\
	\nonumber &= \int_{Z \times X} \log \frac{q^\phi_{Z|x}(z) q_X(x)}{p^\theta_{Z|x}(z) p^\theta_X(x)} d \mathbb{Q}^\phi_{Z \times X}(z,x).
\end{align}
The last of these expressions yields that
\[
\begin{split}
D_{\mathrm{KL}}(\mathbb{Q}^\phi_{Z \times X} \| \mathbb{P}^\theta_{Z \times X}) &= \int_X \log \frac{q_X(x)}{p^\theta_X(x)} d \mathbb{Q}_X(x) 
+ \int_X \int_Z \log \frac{q^\phi_{Z|x}(z)}{p^\theta_{Z|x}(z) } d \mathbb{Q}^\phi_{Z |x}(z) d \mathbb{Q}_X(x) \\
&= D_{\mathrm{KL}}( \mathbb{Q}_X \| \mathbb{P}^\theta_{X} ) + \int_X D_{\mathrm{KL}}( \mathbb{Q}^\phi_{Z | x} \| \mathbb{P}^\theta_{Z|x} ) d \mathbb{Q}_X(x).
\end{split}
\]
The first term in this expression is small when the true distribution $\mathbb{Q}_X$ is hard to distinguish from the distribution of $X$ generated by the decoder $\mathbb{P}^\theta_{X}$. The second term is small when, on average, the conditional distribution of the \emph{encoder} on $Z$ given~$x$ is hard to distinguish from the conditional distribution of the \emph{decoder} on $Z$ given~$x$.

As usual in variational inference (cf. \cite{sci-bib:blei}), we subtract $D_{\mathrm{KL}}(\mathbb{Q}_X \| \mathcal{L}^k)$ and minimize instead
\begin{align}
	\label{sci-eq:lossfn1}
	- D_{\mathrm{KL}}(\mathbb{Q}_X \| \mathcal{L}^k ) &+ D_{\mathrm{KL}}( \mathbb{Q}_X \| \mathbb{P}^\theta_{X} ) + \int_X D_{\mathrm{KL}}( \mathbb{Q}^\phi_{Z | x} \| \mathbb{P}^\theta_{Z|x} ) d \mathbb{Q}_X(x),\\
	\nonumber &= \int_X\left[ -\log q_X(x) + \log \frac {q_X(x)}{p^\theta_{X}(x)}\right] d\mathbb Q_X (x) + \int_X D_{\mathrm{KL}}( \mathbb{Q}^\phi_{Z | x} \| \mathbb{P}^\theta_{Z|x} ) d \mathbb{Q}_X(x) \\
	\nonumber &= -\int_X \log p^\theta_{X}(x) \ d\mathbb Q_X (x) + \int_X D_{\mathrm{KL}}( \mathbb{Q}^\phi_{Z | x} \| \mathbb{P}^\theta_{Z|x} ) d \mathbb{Q}_X(x).
\end{align}
This expression can be recognized as being at the start of the derivation for the loss function used in \cite{sci-bib:vae}.
(We assume $D_{\mathrm{KL}}(\mathbb{Q}_X \| \mathcal{L}^k) < \infty$ and in particular that $\mathbb{Q}_X$ is absolutely continuous with respect to the Lebesgue measure $\mathcal{L}^k$.)

However, the marginal density $p^\theta_X$ is often inaccessible, i.e.~it is often impossible to compute and hard to approximate. Therefore, one rewrites the functional in a different way. By the representation given in~\eqref{sci-eq:KL-after-bayes} we find
\begin{align*}
	D_{\mathrm{KL}}&(\mathbb{Q}^\phi_{Z \times X}\| \mathbb{P}^\theta_{Z \times X}) - D_{\mathrm{KL}}(\mathbb{Q}_X \| \mathcal{L}^k)\\
	&=\int_X \biggl[  \int_Z \log\frac{q^\phi_{Z | x}(z)}{p^\theta_{X|z}(x) p_{Z}(z)}\; d \mathbb Q^\phi_{Z | x} (z) 
	\biggr]  d\mathbb Q_X (x ) \\
	&= - \int_X \int_Z \log p^\theta_{X|z}( x ) d \mathbb{Q}^\phi_{Z | x}(z) d \mathbb{Q}_{X}(x) +
	\int_X \biggl[ \int_Z \log \frac{q^\phi_{Z|x} (z) }{p_{Z}(z)} d \mathbb{Q}^\phi_{Z|x} (z)\biggr] d \mathbb{Q}_{X}(x) 
	\\
	&=  - \int_X \int_{\mathcal{E}} \log p^\theta_{X | \mathsf{Enc}_\phi(x,\epsilon)}(x  )d \mathbb{Q}_\mathcal{E}(\epsilon) d \mathbb{Q}_X(x) + \int_X \int_\mathcal{E} D_{\mathrm{KL}}(\mathbb{Q}^\phi_{Z|x} \| \mathbb{P}_Z) d \mathbb{Q}_\mathcal{E}(\epsilon) d \mathbb{Q}_X(x) \\
	&=\int_{X \times \mathcal{E}} \left[- \log p^\theta_{X | \mathsf{Enc}_\phi(x,\epsilon)}(x) 
	+ 
	D_{\mathrm{KL}}(\mathbb{Q}^\phi_{Z|x} \| \mathbb{P}_{Z} )
	\right] d (\mathbb{Q}_X \otimes \mathbb{Q}_\mathcal{E})(x,\epsilon).
\end{align*}

Our choice of loss function $\mathcal{L}_{\phi,\theta}:X \times \mathcal{E} \to \mathbb{R}$ is therefore
\begin{align}
	\mathcal{L}_{\phi,\theta}(x,\epsilon)
	&:= - \log p^\theta_{X | \mathsf{Enc}_\phi(x,\epsilon)}(x )
	+ D_{\mathrm{KL}} (\mathbb{Q}^\phi_{Z|x} \| \mathbb{P}_{Z} )
	\label{sci-eq:loss-line1} \\
	&= \frac{1}{2 \sigma_X^2} \bigl\| x - \mu_{X,\theta}\bigl(\mathsf{Enc}_\phi(x, \epsilon)\bigr) \bigr\|^2 + \frac{k}{2} \log (2\pi \sigma_X^2) \label{sci-eq:loss-line2}\\
	&\qquad + \frac{1}{2}\left[ \| \mu_{Z,\phi}(x) \|^2 - d
	- \mathrm{tr} \big(\log ( \mathrm{diag}(\sigma_{Z,\phi}(x))^2\big)+  \mathrm{tr}\big(\mathrm{diag}(\sigma_{Z,\phi}(x))^2\big)\right],\nonumber
\end{align}
which up to scaling and a constant agrees with the loss function used in \eqref{sci-eq:loss}.

\medskip	

This derivation allows us to interpret the effects of the different terms and constants in this loss function. The first term in~\eqref{sci-eq:loss-line1} can be interpreted as a (negative) log-likelihood, the probability of observing $x$ conditioned on the property that $z =  \mathsf{Enc}_\phi(x,\epsilon)$. This term is written in detail on the  line~\eqref{sci-eq:loss-line2}, where the Gaussian structure of $p^\theta_{X|z}$ translates into a squared distance weighted by the factor $1/2\sigma_X^2$.

The second term in~\eqref{sci-eq:loss-line1} measures the divergence between the conditional distribution $\mathbb{Q}^\phi_{Z|x}$ and the standard Gaussian. 

For very small values of $\sigma_X$, the first term in~\eqref{sci-eq:loss-line1} dominates the second. In practice, this means that for the parameters $\phi$ and $\theta$ found by the optimization procedure, there is no guarantee that the distribution $\mathbb{Q}^\phi_{Z|x}$ is close to the standard Gaussian measure $\mathbb{P}_Z$; in general it will be far away. Heuristically, the effective range of the encoder will have small $\mathbb{P}_Z$ measure. 

For values of $z$ that are in the effective range of the encoder, the decoder will produce realistic trajectories. However, for the values of $z$ that are not in the range, there is no reason for the decoder to produce realistic trajectories. In particular, the generative model that first independently samples $z \in Z$ and $\omega \in \Omega$ according to $\mathbb{P}_Z$ and $\mathbb{P}_\Omega$ respectively and then computes $\mathsf{Dec}_{\phi}(z,\omega)$, will have very different statistics from the model that samples from $\mathbb{Q}_X$ if $\sigma_X$ is very small.

\bibliographystyle{abbrvnat}
 \bibliography{scibib}
\end{document}